\documentclass[10pt,journal,compsoc]{IEEEtran}

\ifCLASSOPTIONcompsoc
  \usepackage[nocompress]{cite}
\else
  \usepackage{cite}
\fi
\ifCLASSINFOpdf
   \usepackage[pdftex]{graphicx}
   \usepackage{xcolor,tikz, subcaption, color}
\else
\fi
\usepackage{amsmath, amssymb, amsfonts, bm}
\usepackage{bbm}
\usepackage[ruled, vlined, commentsnumbered, linesnumbered]{algorithm2e}
\usepackage{algorithmic}

\usepackage{booktabs,tabularx}
\usepackage[symbol]{footmisc}
\usepackage{array, multirow}

\usepackage{url, multirow, hyperref}
\usepackage{kotex}

\newcommand{\eg}{{\em e.g.}}           % e.g.
\newcommand{\ie}{{\em i.e.}}           % i.e.
\newcommand{\etc}{{\em etc.}}         % etc.

\usepackage{pifont}% http://ctan.org/pkg/pifont
\newcommand{\rev}[1]{\textcolor{black}{#1}}
\newcommand{\ks}[1]{\textcolor{black}{#1}}

\begin{document}
\bstctlcite{IEEEexample:BSTcontrol}
\title{A Quantitatively Interpretable Model for Alzheimer's Disease Prediction using Deep Counterfactuals}

\author{Kwanseok~Oh, 
        Da-Woon~Heo,
        Ahmad~Wisnu~Mulyadi,
        Wonsik~Jung,
        Eunsong~Kang,
        Kun~Ho~Lee,
        and~Heung-Il~Suk,~\IEEEmembership{Senior Member,~IEEE}
        
\IEEEcompsocitemizethanks{
\IEEEcompsocthanksitem K. Oh and D.-W. Heo are with the Department of Artificial Intelligence, Korea University, Seoul 02841, Republic of Korea, e-mail: \{ksohh, daheo\}@korea.ac.kr.
\IEEEcompsocthanksitem A.W. Mulyadi, W. Jung, and E. Kang are with the Department of Brain and Cognitive Engineering, Korea University, Seoul 02841, Republic of Korea, e-mail: \{wisnumulyadi, ssikjeong1, eunsong1210\}@korea.ac.kr.
\IEEEcompsocthanksitem K.H. Lee is with the Department of Biomedical Science and Gwangju Alzheimer's \& Related Dementia Cohort Research Center, Chosun University, Gwangju 61452, Republic of Korea, and Korea Brain Research Institute, Daegu 41062, Republic of Korea, e-mail: leekho@chosun.ac.kr.
\IEEEcompsocthanksitem H.-I. Suk is with the Department of Artificial Intelligence and the Department of Brain and Cognitive Engineering, Korea University, Seoul 02841, Republic of Korea, e-mail: hisuk@korea.ac.kr.
\IEEEcompsocthanksitem H.-I. Suk and K.H. Lee are the co-corresponding authors.
% \IEEEcompsocthanksitem Co-corresponding authors: hisuk@korea.ac.kr (Heung-Il Suk) and leekho@chosun.ac.kr (Kun Ho Lee)
}}

% \author{Michael~Shell,~\IEEEmembership{Member,~IEEE,}
%         John~Doe,~\IEEEmembership{Fellow,~OSA,}
%         and~Jane~Doe,~\IEEEmembership{Life~Fellow,~IEEE}% <-this % stops a space
% \IEEEcompsocitemizethanks{\IEEEcompsocthanksitem M. Shell was with the Department
% of Electrical and Computer Engineering, Georgia Institute of Technology, Atlanta,
% GA, 30332.\protect\\
% % note need leading \protect in front of \\ to get a newline within \thanks as
% % \\ is fragile and will error, could use \hfil\break instead.
% E-mail: see http://www.michaelshell.org/contact.html
% \IEEEcompsocthanksitem J. Doe and J. Doe are with Anonymous University.}% <-this % stops an unwanted space
% \thanks{Manuscript received April 19, 2005; revised August 26, 2015.}}

% \markboth{IEEE TRANSACTIONS ON PATTERN ANALYSIS AND MACHINE INTELLIGENCE}%
\markboth{PREPRINT}%
{Oh \MakeLowercase{\textit{et al.}}: \MakeUppercase{A Quantitatively Interpretable Model for Alzheimer's Disease Prediction using Deep Counterfactuals}}

\IEEEtitleabstractindextext{%
\begin{abstract}
Deep learning (DL) for predicting Alzheimer's disease (AD) has provided timely intervention in disease progression yet still demands attentive interpretability to explain how their DL models make definitive decisions. Recently, $\emph{counterfactual reasoning}$ has gained increasing attention in medical research because of its ability to provide a refined visual explanatory map. However, such visual explanatory maps based on visual inspection alone are insufficient unless we intuitively demonstrate their medical or neuroscientific validity via quantitative features. In this study, we synthesize the counterfactual-labeled structural MRIs using our proposed framework and transform it into a gray matter density map to measure its volumetric changes over the parcellated region of interest (ROI). We also devised a lightweight linear classifier to boost the effectiveness of constructed ROIs, promoted quantitative interpretation, and achieved comparable predictive performance to DL methods. Throughout this, our framework produces an ``\emph{AD-relatedness index}'' for each ROI and offers an intuitive understanding of brain status for an individual patient and across patient groups with respect to AD progression.
\end{abstract}

% Note that keywords are not normally used for peer review papers.
}

\maketitle

\IEEEdisplaynontitleabstractindextext
\IEEEpeerreviewmaketitle

\IEEEraisesectionheading{\section{Introduction}\label{sec:introduction}}
    \IEEEPARstart{A}{lzheimer’s} disease (AD) is a neurodegenerative brain disease characterized by memory loss, logical thinking difficulties, speech impairment, and problems with reading and writing~\cite{alzheimer20192019}. Although many efforts have been made to enhance understanding and discover efficient treatment, the medication for AD is intended only to slow AD progression~\cite{weiner2013alzheimer}. As numerous studies have revealed the advantages of early intervention~\cite{cummings2007disease}, it is paramount to identify patients who have mild cognitive impairment (MCI)—a prodromal stage of AD that is likely to convert to AD~\cite{petersen2009mild}—to effectively delay cognitive decline. With advances in brain imaging techniques, such as structural magnetic resonance imaging (sMRI), a predictive framework that integrates clinical examination and imaging techniques~\cite{schneider2009neuropathology} has been developed to understand brain disease-related structural changes. However, clinical examinations may vary contingent on which diagnosis index is used, as each includes various evaluation items. Thus, the visual inspection of such brain imaging scans inevitably depends on the qualitative evaluations and subjective decisions of radiologists and clinicians.
    
    Originally derived from oncological studies~\cite{aerts2014decoding,yu2016predicting}, \textit{radiomics} has recently been used to extract quantitative features from brain imaging data to present objective and reliable disease-related characteristics and reflect anatomical landmarks. These quantitative features (\eg, density, volume, morphometry, and textures) have phenotypic characteristics beneficial for disease analysis~\cite{gillies2016radiomics}. Concurrently, quantitative feature-based machine learning (ML) techniques have been proposed as an essential component of computer-aided diagnosis and detection systems~\cite{lopez2009svm}. Unlike statistical methods~\cite{zhou2019dual,lee2020magnetic}, which are based on group-level analysis, ML allows individual-level precise predictions for subjects~\cite{ij2018statistics} to recognize the intricate patterns of input features for downstream tasks. Although existing ML-based methods help identify MCI/AD at the individual-level, most of the biomarker region selection is biased toward a small set of pre-defined regions for the sake of validation alone~\cite{li2019stability}. That is, undiscovered disease-related regions are ignored, the regions that could feasibly be considered potential biomarkers. Moreover, obtaining sufficient reproducibility and generalizability over regions of interest (ROIs) used for assessment is challenging because the disease-relevant regions are manually predetermined~\cite{zhao2020independent,lee2020magnetic}.

    \begin{figure*}[t]
        \centering
        \subfloat[Counterfactual-labeled dataset generation. According to a series of preprocessing, we obtained the \rev{r-sMRIs $\mathbf{X}^r$} from raw brain images and trained the predictive model for AD classification, using the \rev{r-sMRIs $\mathbf{X}^r$} as input. By performing counterfactual reasoning via the pre-trained classifier, we synthesize the \rev{c-sMRIs $\mathbf{X}^c$} conditioned on target labels.]{\label{fig:cf-gen}{\includegraphics[width=.88\linewidth ]{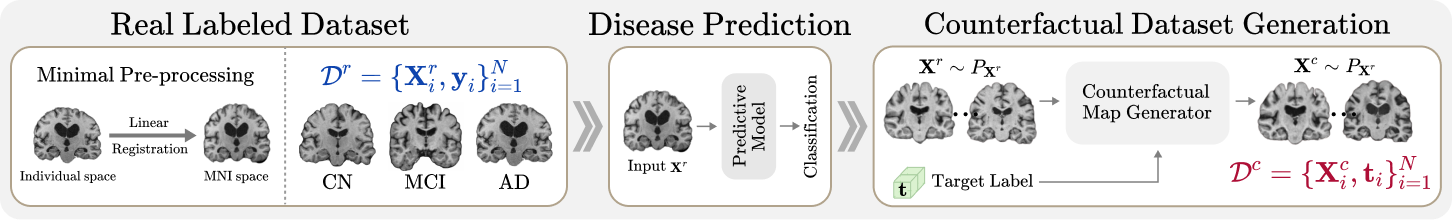}}}
        \vspace{0.1cm}
        \subfloat[AD-effect map estimation for compositing \rev{AD-effect ROIs $\hat{\mathbf{x}}_q$}. Given the \rev{r-sMRIs $\mathbf{X}^r$} and their counterparts \rev{c-sMRIs $\mathbf{X}^c$}, those images are manipulated as GM density maps  (\ie, rGM and cGM) to perform the numerical measurement for the quantitative explanation of derived counterfactual reasoning. If we suppose CN as true (rGM), it indicates that cMCI or cAD is a counterfactual (cGM) corresponding to an alternative scenario.]{\label{fig:ad-effect-map}{\includegraphics[width=.88\linewidth ]{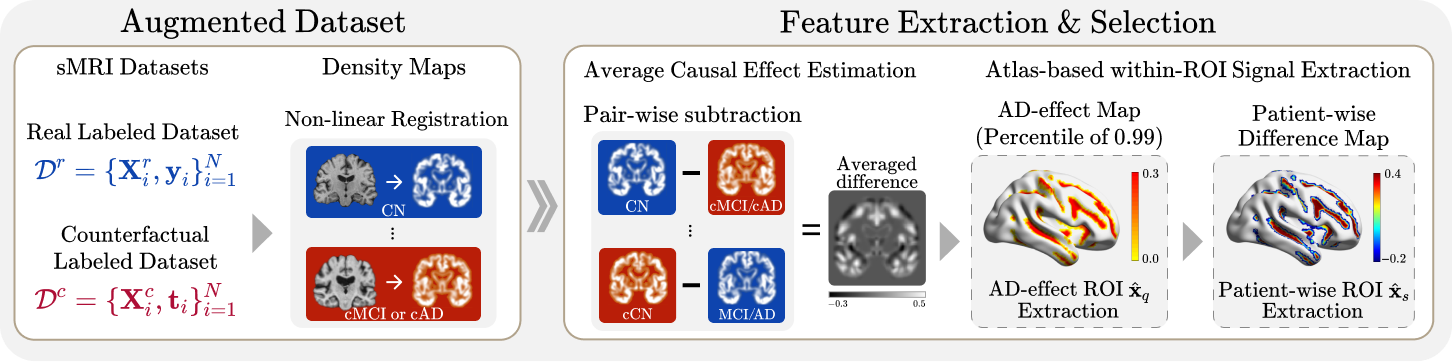}}}
        \vspace{0.1cm}
        \subfloat[A schematic overview of our LiCoL framework. The gray boxes denote the embedding layer. ``+'' and ``$\otimes$'' denote element-wise addition and element-wise multiplication, respectively. The query is fixed as AD-effect ROIs $\hat{\mathbf{x}}_q$; whereas the key and value are individually initialized to the patient-wise ROIs $\hat{\mathbf{x}}_s$.]{\label{fig:licol}{\includegraphics[width=.88\linewidth ]{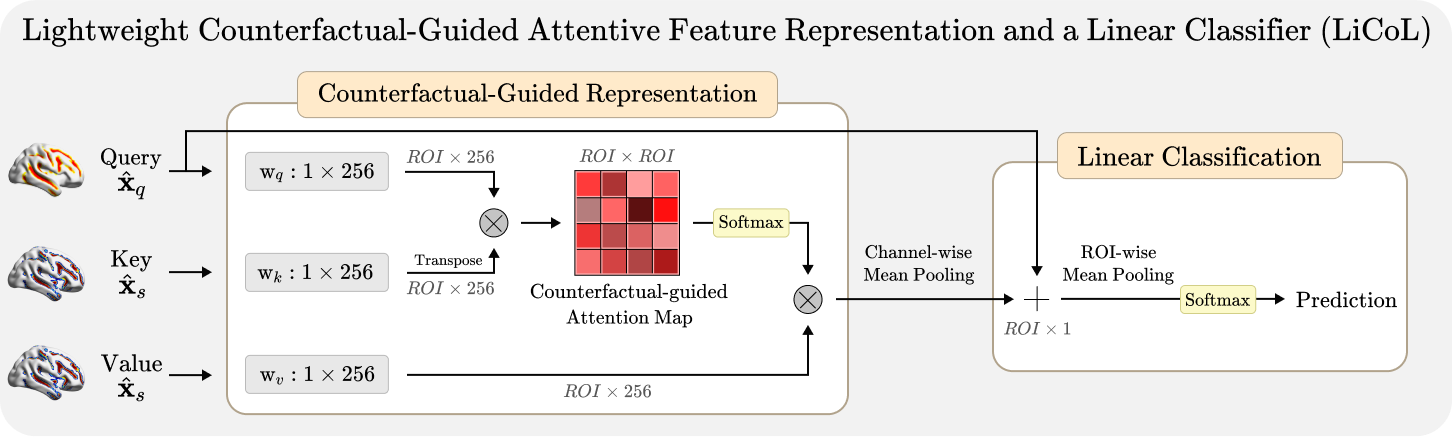}}}
        \caption{Illustration of our gray matter \rev{(GM)} density-based analysis workflow.}
    \label{fig:overall-framework}
    \end{figure*}
    
    Deep learning (DL)~\cite{lecun2015deep} techniques have recently achieved impressive outcomes in the medical field and have made remarkable progress in 3D sMRI-based AD prediction than the conventional ML methods~\cite{venugopalan2021multimodal,zhang2021explainable,yang2021deep}. Having devised with a hierarchical structure, DL models are proficient in discovering informative patterns in a data-driven manner, which alleviates the need for an a priori handcrafted selection of disease-related regions by experts. Despite DL models significantly enhancing predictive performance, providing the underlying interpretable final decisions remains a longstanding goal. This unfavorable ``black-box'' nature of DL models impedes many studies from clearly explaining and interpreting how their proposed DL models make definitive decisions~\cite{adadi2018peeking}. To resolve this issue, visual explanation-based approaches in the field of explainable artificial intelligence (XAI)~\cite{arrieta2020explainable} have increasingly been exploited for interpreting DL models for medical image analysis. Existing methods, such as attribution-based approaches~\cite{sundararajan2017axiomatic,selvaraju2017grad}, offer the visual saliency map as explanatory evidence by analyzing the gradients or activations of the model. While such visual saliency maps can highlight the influences and contributions of the class-relevant features (\eg, particular disease-related ROIs) \emph{w.r.t} the prediction in given input images, they tend to produce similar saliency maps across the wrong class labels, owing to an ``attribution vanishing'' problem~\cite{wang2019learning}.

    Counterfactual reasoning has gradually emerged as an alternative because it can provide refined visual explanatory maps, called counterfactual maps (CF maps), which exhibit fundamental explanations regarding the model's decisions in various hypothetical scenarios, similar to the human decision-making process~\cite{byrne2019counterfactuals}. Specifically, a CF map enables an observer to consider contrastive explanations and causal inference, such as why a particular decision is made instead of another, as well as observe how an alteration of specific attributes affects the model's output~\cite{goyal2019counterfactual}. Most preceding studies that yielded counterfactual reasoning in DL were built upon the generative adversarial network (GAN)~\cite{goodfellow2016nips} and its variants~\cite{mirza2014conditional,dash2022evaluating,oh2022learn}, thereby it is proficient at generating a potential outcome that reveals unobservable instances. In terms of MCI/AD prognosis, the generated CF map serves as a visual explanatory map that includes anatomical changes such as brain hypertrophies or atrophies, assuming that the cognitively normal (CN) is diagnosed as a patient (refer to Fig.~\ref{fig:cf-map-result}).

    In practice, the shortcoming of such CF maps is that they still rely on visual inspection according to voxel-level morphometry such that these visual explanations are merely synthesized on the original images to understand the model's decision. Similarly, most XAI-inferred medical imaging studies~\cite{oh2022learn,xia2021learning,zhang2021explainable} generally provided a visual interpretation of predictive performance without properly proving their medical or neuroscientific validity. This issue limits the yielding of intuitive insights from a clinical perspective~\cite{kim2019design}. In practical application, exhibiting explanatory maps alone is not self-sufficient, nor do they fulfill an expectation to support MCI/AD prediction as clinically decisive auxiliary information. \textit{Thus, a key motivation for this study is to advance beyond the limitations of the DL-based visual explanatory in medical image analysis by making the insights derived from such methodologies more measurable, intuitive, and scalable to healthcare professionals, including radiologists and clinicians.}
    % it is vital to intuitively demonstrate the validity of their visual attribution by analyzing the counterfactual reasoning via quantitative features~\cite{schwarz2016large}.

    For these premises, we propose a novel framework for counterfactual-induced feature representation and quantitatively interpretable and explainable AD identification. As an extension of our prior study, we exploit LEAR's counterfactual map generator (CMG)~\cite{oh2022learn} to synthesize the counterfactual-labeled sMRI (c-sMRI) using the generated CF map via the given input (\ie, real-labeled sMRI, namely, r-sMRI), as depicted in Fig.~\ref{fig:cf-gen}. The newly produced c-sMRIs reflect the target counterfactual attributes with high confidence, so we employed the c-sMRIs as a disease-associated data augmentation set.
    % The newly produced c-sMRIs reflect the target counterfactual attributes with high confidence. Hence, we deduced the constructed c-sMRIs as a disease-associated data augmentation set. 
    By utilizing data-driven knowledge as augmented data, we can unlock a way to investigate the brain that has not yet been converted to the disease-manifested brain or even a clinically improbable case of reversed disease progression by producing alternative scenarios. We then transform both r-sMRIs and c-sMRIs into a gray matter (GM) density map (\ie, rGM and cGM) to precisely measure the volumetric changes in GM~\cite{zhang2016bayesian}, associated with brain atrophies by aging~\cite{xia2021learning} and AD progression~\cite{hirata2005voxel,karas2004global}.
    
    Given the rGM and cGM, we perform a series of GM density-based analyses (Fig.~\ref{fig:ad-effect-map}) and assessments (Fig.~\ref{fig:licol}). First, we obtain the representative difference map by subtracting rGMs from their counterpart cGMs, followed by taking the first statistical moment (\ie, mean of difference maps). In this way, we estimate the so-called ``\emph{average causal disease effect}'' (ACE) approximated by the difference in outcomes between the true (rGM) and the alternative scenario (cGM) from the same subjects. We refer to an estimated ACE as an \emph{AD-effect map}, which portrays certain regions over the AD-affected anatomical variations within the AD spectrum. Subsequently, we explore the significant ROIs (\ie, AD-effect ROIs) by extracting prominent regions from the AD-effect map, such as the feature selection strategy. 
    % We then devise an ROI-based analysis mechanism that grants region-wise investigation for the quantity of volumetric change \emph{w.r.t} the GM density. 
    We further develop a shallow network of lightweight counterfactual-guided attentive feature representation and a linear classifier (LiCoL). Interestingly, the internal operation of the LiCoL from the input to the output can be rewritten as a linear function, as revealed in Supplementary H. This process helps interpret the counterfactual-guided attentive representation as the regional status of a brain, called an ``\emph{AD-relatedness index}'', and explains the output decision quantitatively based on the regional AD-relatedness index~\cite{davatzikos2009longitudinal}. To our best knowledge, this work is the first sMRI-based AD prediction model that makes an input observation interpretable as a quantitative AD-relatedness index according to a DL-based visual explanation. 
    % With this, our method is able to provide the numerical quantity of AD-related landmarks of a single subject and across subject groups. 
    Concretely, the main contributions of our study are summarized as follows:
    \begin{itemize}
        \item We propose a novel methodology\footnote{Publicly available at \url{https://github.com/ku-milab/LiCoL}} to develop fundamental scientific insights from a counterfactual reasoning-based explainable learning method. We demonstrate that our proposed method can interpret intuitively from the clinician's perspective by converting counterfactual-guided deep features to the quantitative volumetric feature domain rather than directly inspecting DL-based visual attributions.
    
        \item We achieved similar or better performance than DL models by designing a LiCoL with the AD-effect ROIs considered to be the distinctive AD-related landmarks via counterfactual-guided deep features.
        
        \item By exploiting our proposed LiCoL, we provide a numerically interpretable AD-relatedness index for each patient as well as patient groups \emph{w.r.t} anatomical variations caused by AD progression.
        
        \item We further investigated the discoveries of AD-manifested subregions based on common and potential regions by comparing the Alzheimer's Disease Neuroimaging Initiative (ADNI) dataset~\cite{MUELLER2005869} and the Gwangju Alzheimer's and Related Dementia (GARD) cohort dataset~\cite{choi2019apoe}.
    \end{itemize}

    \section{Method}\label{sec:methods}
    % In Fig.~\ref{fig:overall-framework}, we depict the principal ideas of our framework, which is comprised of three subsequent crucial streams: (i) the counterfactual-labeled dataset generation, (ii) the AD-effect map estimation for compositing AD-effect ROIs, and (iii) the investigation of ROI significance using our LiCoL.
    
    \subsection{Counterfactual visual explanation model}
    In diagnosing clinical stages across the AD spectrum (\ie, CN/MCI/AD), the goal of the counterfactual visual explanation approach proposed in our LEAR~\cite{oh2022learn} is to produce counterfactual reasoning for the decision of a predictive model $\mathcal{C}$ (Fig.~\ref{fig:cf-gen}). It consists of several key components: the counterfactual map generator (CMG), reasoning evaluator (RE), and discriminator (DC). Specifically, the CMG generates a CF map conditioned on an arbitrary target label, whereas the RE effectively guides the CMG toward comprehending the target label attributes in synthesizing realistic desired images. By exploiting the DC for carefully appraising the real and synthesized images (\ie, r-sMRIs $\mathbf{X}^r$ and c-sMRIs $\mathbf{X}^c$), the CMG will ascertain that the c-sMRIs are constrained as being realistic conversions. It should be noted that the RE is defined as the pre-trained predictive model $\mathcal{C}$, and the structure of the DC is identically imitated in the predictive model $\mathcal{C}$. Initially, we conducted pre-training for the predictive model $\mathcal{C}$ (see Supplementary B) using supervised learning with training samples $\mathbf{X}^r$ and one-hot encoded ground-truth labels $\mathbf{y}$:
    \begin{equation}
        \mathcal{L}^{\mathcal{C}}_{\text{cls}} = \mathbb{E}_{\mathbf{X}^r\sim P_{\mathbf{X}^r}}\left[\operatorname{CE}\left(\mathcal{C}(\mathbf{X}^r),\mathbf{y}\right)\right],\label{predictive_cls_loss}
    \end{equation}
    where $\operatorname{CE}$ is a cross-entropy function. Henceforth, during the subsequent training phase to generate the CF maps, we fixed the weights of the RE while jointly tuning the trainable parameters of DC $\mathcal{D}_\psi$ with the generator $\mathcal{G}_{\phi}$ in CMG.
    
    \subsubsection{CMG architecture}
    The CMG is a variant of a conditional GAN~\cite{mirza2014conditional} devised to effectively synthesize a CF map conditioned on a target label $\mathbf{t}$, where $\mathbf{t}\in[0,1]^{\lvert\mathcal{Y}\lvert}$ with $\lvert\mathcal{Y}\lvert$ denotes the size of the class distribution $\mathcal{Y}$. Suppose that the c-sMRI can be induced through $\mathbf{X}^c=\mathbf{X}^r+\mathbf{M}_{\mathbf{X}^r,\mathbf{t}}$ given an input $\mathbf{X}^r$, along with a target-specific CF map $\mathbf{M}_{\mathbf{X}^r,\mathbf{t}}$. The CMG shall be optimized in generating these maps $\mathbf{M}_{\mathbf{X}^r,\mathbf{t}}$ such that the c-sMRI $\mathbf{X}^c$ is diagnosed as the target label $\mathbf{t}$ with high confidence. Specifically, the architecture of the CMG comprises the encoder $\mathcal{E}_{\theta}$ and the generator $\mathcal{G}_{\phi}$, that is, a variant of U-Net~\cite{ronneberger2015u} with a target label $\mathbf{t}$ concatenated to the skip connections. 
    % The subscripts $\theta$ and $\phi$ of the encoder $\mathcal{E}_{\theta}$ and generator $\mathcal{G}_{\phi}$ indicate the trainable parameters of the respective networks. 
    % Note that we replicate $\mathcal{E}_{\theta}$ from the pre-trained predictive model $\mathcal{C}$ in both its architecture and weights (fixed). Thus, the encoder $\mathcal{E}_{\theta}$ is adequate for extracting a meaningful representation that reflects the class-relevant discriminative features from an input r-sMRI $\mathbf{X}^r$.
    Note that we replicate $\mathcal{E}_{\theta}$ from the pre-trained predictive model $\mathcal{C}$ in both its architecture and weights (fixed) so that $\mathcal{E}_{\theta}$ adequately extracts the class-relevant features from $\mathbf{X}^r$.
	 
    As the CMG should consider the target label $\mathbf{t}$ to produce the CF map, we fuse the target-specific characteristics onto the feature maps obtained from the encoder $\mathcal{E}_{\theta}(\mathbf{X}^r)$ via a concatenation operation. For this purpose, we tile the target label to match the shape corresponding to the respective feature maps of the $l$-th convolution layer. 
    % The size of the tiled target label is $w_l\times h_l\times d_l\times c_l$, where $w_{l}$, $h_{l}$, and $d_{l}$ denote, respectively, the width, height, and depth of a feature map from the $l$-th convolution block, and $c_l$ denotes the number of channels. 
    To obtain the hierarchical discriminative representations \emph{w.r.t} the target label, we devise an additional module that comprises a convolution operation (Conv3D) with a trainable $3\times3\times3$ kernel, a stride of one in each dimension, and zero padding, followed by a Leaky-ReLU activation function ($\operatorname{LReLU}$):
    \begin{equation}
        \tau(\mathbf{F}_{l}^{\mathcal{E}_{\theta}(\mathbf{X}^r)},\mathbf{t}) = \operatorname{LReLU}\left(\operatorname{Conv3D}\left(\mathbf{F}_{l}^{\mathcal{E}_{\theta}(\mathbf{X}^r)}\oplus\operatorname{Tile}(\mathbf{t})\right)\right),
    \end{equation}
    where $\oplus$ denotes an operator of channel-wise concatenation, and $\{\mathbf{F}_l^{\mathcal{E}_{\theta}(\mathbf{X}^r)} \}^L_{l=1}$ denote the output feature maps of the $L$ convolution layers from the encoder $\mathcal{E}_{\theta}(\mathbf{X}^r)$. Subsequently, the target-fused feature maps $\tau(\mathbf{F}_{l}^{\mathcal{E}_{\theta}(\mathbf{X}^r)},\mathbf{t})$ are transferred to the generator $\mathcal{G}_{\phi}$ via skip connections. The generator $\mathcal{G}_{\phi}$ is then capable of seamlessly generating a CF map $\mathbf{M}_{\mathbf{X}^r,\mathbf{t}}$ from the target label-informed feature maps:
        \begin{equation}
            \mathbf{M}_{\mathbf{X}^r,\mathbf{t}}=\mathcal{G}_\phi\left(\mathcal{T}(\mathbf{X}^r,\mathbf{t})\right),
        \end{equation}    
    where $\mathcal{T}(\mathbf{X}^r,\mathbf{t})=\{\tau(\mathbf{F}_{1}^{\mathcal{E}_{\theta}(\mathbf{X}^r)},\mathbf{t}),...,\tau(\mathbf{F}_{L}^{\mathcal{E}_{\theta}(\mathbf{X}^r)},\mathbf{t})\}$. Lastly, we produce a c-sMRI by incorporating the CF map with an input $\mathbf{X}^r$ using addition, \ie, $\mathbf{X}^c=\mathbf{X}^r+\mathbf{M}_{\mathbf{X}^r, \mathbf{t}}$, which is supposed to be diagnosed as the target label $\mathbf{t}$.
    
    \subsubsection{The objective function of the CMG}
    % This subsection thoroughly presents the essential objective function for optimizing the overall counterfactual visual explanation framework. Here, we effectively generate a refined explanatory map using an adversarial learning strategy in conjunction with several auxiliary losses.
    To generate a realistic c-sMRI $\mathbf{X}^c$, we adopt the least square GAN (LSGAN)~\cite{mao2017least} loss function to enforce a stable optimization by penalizing samples far from the DC's decision boundary. Guided by this loss, the DC assists the CMG such that it will thoughtfully minimize the substantial distance between the real and generated distributions:
	\begin{align}
    	\begin{split}
    	 \mathcal{L}_{\text{adv}}^{\mathcal{D}_{\psi}} &= \mathbb{E}_{\mathbf{X}^r\sim P_{\mathbf{X}^r}}\left[(\mathcal{D}_{\psi}(\mathbf{X}^r)-1)^2\right]\\ &+ \frac{1}{2}\left(\mathbb{E}_{\mathbf{X}^r\sim P_{\mathbf{X}^r}}\left[\mathcal{D}_{\psi}(\mathbf{X}^c)^2+\mathcal{D}_{\psi}(\tilde{\mathbf{X}}^{r})^2\right]\right),
    	 \end{split} \label{adv d loss}
    	\\
    	\begin{split}
    	\mathcal{L}_{\text{adv}}^{\mathcal{G}_{\phi}} &= \frac{1}{2}\left(\mathbb{E}_{\mathbf{X}^r\sim P_{\mathbf{X}^r}}\left[(\mathcal{D}_{\psi}(\mathbf{X}^c) - 1)^2+(\mathcal{D}_{\psi}(\tilde{\mathbf{X}}^r) - 1)^2\right]\right),
            \end{split}\label{adv g loss}
    \end{align}
    where $\mathbf{X}^c$ and $\tilde{\mathbf{X}}^r = \mathbf{X}^c+\mathbf{M}_{\mathbf{X}^c,\mathbf{y}^r}$ denote the c-sMRI and its respective override c-sMRI (will also be utilized for cycle consistency), respectively, and $P_{\mathbf{X}^r}$ denotes the distribution of r-sMRI samples. Note that $\mathbf{M}_{\mathbf{X}^c,\mathbf{y}^r}$ indicates the CF map overriding the c-sMRI $\mathbf{X}^c$ to obtain $\tilde{\mathbf{X}}^r$ which is expectedly to resemble the r-sMRI $\mathbf{X}^r$ conditioned by $\mathbf{y}^r=\mathcal{C}(\mathbf{X}^r)$.
    
    Because the DC is employed merely to differentiate between the input $\mathbf{X}^r$ and the generated $\mathbf{X}^c$ and $\tilde{\mathbf{X}}^r$, it does not have the immediate capacity to explicitly guide the CMG to either preserve the appearance of the input or endorse the target attribution during the generative process. Thus, we employ a cycle consistency loss~\cite{zhu2017unpaired} based on the $\ell_1$-norm to produce enhanced target-dependent CF maps. In this way, we force the CMG to consistently strive for various conditions without suffering from a mode collapse~\cite{goodfellow2016nips}:
    \begin{equation}
        \begin{aligned}
            \mathcal{L}_{\text{cyc}} = \mathbb{E}_{\mathbf{X}^r\sim P_{\mathbf{X}^r},\mathbf{t}\sim U(0,\lvert\mathcal{Y}\lvert)}\left\|\tilde{\mathbf{X}}^r - \mathbf{X}^r\right\|_1,
        \end{aligned}\label{eq_cyc}
    \end{equation}
    where $\mathbf{t}\sim U(0,\lvert\mathcal{Y}\lvert)$ denotes the one-hot encoded vector drawn randomly from a discrete uniform distribution.

	Furthermore, we utilize the total variation loss to reconcile the elaborate synthesis between the input $\mathbf{X}^r$ and its CF map $\mathbf{M}_{\mathbf{X}^r, \mathbf{t}}$. In particular, total variation loss ensures that the c-sMRI $\mathbf{X}^c$ imposes local spatial continuity and smoothness to mitigate unnatural and overly pixelated results, which encourages visual coherence in a real image:
    \begin{align}
    \begin{split}
        	\mathcal{L}_{\text{tv}} = \sum_{i,j,k}\left\lvert\mathbf{X}^c_{i+1,j,k}-\mathbf{X}^c_{i,j,k}\right\lvert &+ \left\lvert\mathbf{X}^c_{i,j+1,k}-\mathbf{X}^c_{i,j,k}\right\lvert\\ &+ \left\lvert\mathbf{X}^c_{i,j,k+1}-\mathbf{X}^c_{i,j,k}\right\lvert,
    \end{split}\label{eq_tv}
    \end{align}
    where $\mathbf{X}^c = \mathbf{X}^r + \mathbf{M}_{\mathbf{X}^r, \mathbf{t}}$ and $i$, $j$, and $k$ are the 3D coordinates of each index in volumetric images.
    
    Elastic regularization is also applied to the CMG to impose a constraint on the magnitude of the CF map $\mathbf{M}_{\mathbf{X}^r, \mathbf{t}}$. By doing this, the vast majority of fine-grained feature attributions for counterfactual reasoning are highlighted such that minimal modifications over the features induce a converted prediction from $\mathbf{y}^r=\mathcal{C}(\mathbf{X}^r)$ to the target $\mathbf{t}$.
    \begin{equation}
        \mathcal{L}_{\text{map}} = {\mathbb{E}_{\mathbf{X}^r\sim P_{\mathbf{X}^r}}}\left[\lambda_1\left\|\mathbf{M}_{\mathbf{X}^r, \mathbf{t}}\right\|_1 + \lambda_2\left\|\mathbf{M}_{\mathbf{X}^r, \textbf{t}}\right\|_2\right],\label{cm loss}
    \end{equation}
    where $\lambda_1$ and $\lambda_2$ are weighting constants.
    
    As adequate assistance to the generator $\mathcal{G}_\phi$, we employ the classification loss function to produce CF maps that transform an input $\mathbf{X}^r$ so that it is precisely classified as a target label $\mathbf{t}$:
    \begin{equation}
        \mathcal{L}_{\text{cls}} = \mathbb{E}_{\mathbf{X}^c\sim P_{\mathbf{X}^r}}\left[\operatorname{CE}\left(\mathcal{C}(\mathbf{X}^c), \mathbf{t})\right)\right].\label{cls_loss}
    \end{equation}
    
    Consequently, we define the composite objective function for the overall CF map generation task as follows:
    \begin{equation}
\mathcal{L}_{\text{CMG}}=\lambda_{3}\mathcal{L}^{\mathcal{G}_\phi}_{\text{adv}}+\lambda_{4}\mathcal{L}^{\mathcal{D}_\psi}_{\text{adv}}+\lambda_{5}\mathcal{L}_{\text{cyc}}+\lambda_{6}\mathcal{L}_{\text{cls}}+\lambda_{7}\mathcal{L}_{\text{tv}}+\mathcal{L}_{\text{map}},\label{total_cmg_loss}
	\end{equation}
    where $\lambda_{*}$ values are the hyperparameters for model training. We empirically tune $\lambda$ such that the magnitudes of the gradients for each loss term are roughly balanced.

    \subsection{Quantifying explainability of counterfactual-guided deep features}
    Having sufficiently optimized the CF map generation task, we extend our work~\cite{oh2022learn} in quantifying the explainability of the acquired counterfactual-guided deep features in terms of AD prediction. For this purpose, we carefully devise the overall explainability quantification steps, including (i) establishment of image manipulation, (ii) extraction of AD-related landmarks, and finally (iii) usage of a lightweight counterfactual-guided attentive feature representation and a linear classifier (LiCoL). The procedures of this in-depth analysis are depicted in Fig.~\ref{fig:ad-effect-map} and Fig.~\ref{fig:licol}.
	
    \subsubsection{Establishment of image manipulation}\label{sec: manipulation}
    This brain image manipulation step aims to establish a proportional approach to quantify the explainability of deep features as a gray matter (GM) density map. The utilization of a GM density map assists in conducting accurate numerical measurements and quantitative feature-based analysis of brain disease. To this end, we devise subsequent procedures to reverse the process used for acquiring the train-ready input $\mathbf{X}^r$ (\ie, r-sMRIs) from a raw brain image $\mathbf{B}$. In this way, the GM density map, including counterfactual-guided deep features, is obtained by manipulating the reflective superimposed processing.

    Specifically, both r-sMRI $\mathbf{X}^r$ and the corresponding c-sMRI $\mathbf{X}^c$ are fed into a series of reverse preprocessing steps consisting of reverse Gaussian normalization, reverse quantile normalization, and up-scaling steps. To calculate the reverse Gaussian normalization, each mean and standard deviation of the subject is stored during preprocessing and reused to recover the original values. The subsequent step is reverse quantile normalization; however, it is intractable to process the c-sMRI $\mathbf{X}^c$ by simply reversing the quantile normalization, as opposed to r-sMRI $\mathbf{X}^r$. Since the previous values of voxels where normalization was performed by quantile thresholding are unknown after synthesizing with a CF map, we resort to a histogram-matching technique as a surrogate solution. By viewing the model's input as an alternative ground truth, which is paired with the $\mathbf{X}^c$ being processed, we adequately circumvent this issue through the alignment of quantile-normalized voxels. Thereafter, the final reversed raw image $\mathbf{B}'$ is acquired using an up-scaling operation to match the size of $\mathbf{B}$ (\ie, $193\times229\times193$). 

    Furthermore, a few auxiliary steps are carried out to produce the GM density map over a $\mathbf{B}'$ set. There are four kinds of steps: (i) as the $\mathbf{B}'$ was skull-stripped and linearly registered into MNI152 space, those images are instantly segmented into GM, white matter, and cerebrospinal fluid volume probability maps using FMRIB's automated segmentation tool~\cite{zhang2001segmentation}, (ii) segmented GM images are nonlinearly registered using FMRIB’s nonlinear registration tool to generate GM density maps for each image, (iii) we then modulated these GM density maps via Jacobian of the warp field, (iv) the resulting GM density maps are finally smoothed with an isotropic Gaussian kernel with a $\sigma$ of 2 mm for alleviating contrast and other irrelevant details. It should be noted that values in acquired GM density maps refer to the intensity of GM density within the brain regions.
    	
    \subsubsection{Extraction of AD-related landmarks}\label{sec: AD-related landmarks}
    Given a set of manipulated GM density maps, we further investigate the validity of counterfactual-guided deep features, applying the average causal disease effect for GM density-based analysis. Let ${\hat{\mathbf{X}}}^{r}$ and $\hat{\mathbf{X}}^{c}$ denote the rGM and the cGM manipulated via an r-sMRI $\mathbf{X}^r$ and the c-sMRI $\mathbf{X}^c$, respectively. We first subtract rGMs $\hat{\mathbf{X}}^r$ from their corresponding cGMs $\hat{\mathbf{X}}^c$ in diverse clinical stages to acquire the stage-wise difference maps. For instance, for a CN-labeled rGM, the counterpart of its clinical label for the cGM shall be defined as the AD or MCI, depending on the given scenario. The favorable properties of difference maps encompass a discriminative capability to prominently reflect inherent disease-associated regional localization as well as the unique anatomical characteristics between the different disease-state groups. These difference maps are then averaged into a representative difference map, which refers to as the AD-effect map $\mathbf{M}_{\text{eff}}$, by exploiting the first moment in statistics (\ie, the mean of difference maps) as:
        \begin{equation}
            \mathbf{M}_{\text{eff}} = \frac{1}{N} \sum^N_{i=1}\left\lvert\hat{\mathbf{X}}^{r}_{i} - \hat{\mathbf{X}}^{c}_{i}\right\lvert,\label{eq:AD-effect-map}
        \end{equation}
        where $N$ and $\lvert\cdot\lvert$ denote the total number of samples and the absolute operation, respectively. Note that the purpose of applying an absolute operation is to emphasize areas where the magnitude of GM density alterations between clinical stage groups are prominent, considering the variability of AD-sensitive regions. Such an AD-effect map $\mathbf{M}_{\text{eff}}$ is finally masked by the percentile threshold that reveals the most distinctly highlighted regions among the set of fine-grained regions (illustrated in Fig.~\ref{fig:ad-effect-map}).
        
    \subsubsection{AD-effect ROI composition}\label{sec:ad-effect-roi-composition}
    Brain parcellation not only provides an understanding of the fundamental brain organization and function of closely interacting regions but also compresses information from hundreds of thousands of voxels or vertices into manageable sets. Under these characteristics, we adopt an ROI-based analysis that extracts and analyzes significant ROIs across voxel-level values, which are residing on the AD-effect map $\mathbf{M}_{\text{eff}}$. To this end, the automated anatomical labeling (AAL3) atlas~\cite{rolls2020automated} (see Supplementary G) is overlaid on the AD-effect map to further subdivide it into individual regions, so that each trimmed region is completely within the cortical and subcortical regions and has anatomical specificity. We utilize a set of voxel indices $\mathcal{V}$ across the parcellated regions to aggregate the region-wise highlighted voxels on the AD-effect map with a thresholding parameter $\alpha$ (see Supplementary C and Supplementary J). We then take the average over the total number of highlighted voxels in each region. Eventually, representative ROIs are defined by concatenating the region-wise aggregated values as:
    
    \begin{equation}\label{eq:ad-effect-roi}
        \hat{\mathbf{x}}_q := \bigg\Vert_{r=1}^R\left(\frac{1}{\# \mathcal{V}_r}\sum^{\# \mathcal{V}_r}_{i=1}  \mathbf{M}_{\text{eff}}[\mathcal{V}_r[i]] \ * \ \mathbbm{1}(\# \mathcal{V}_r \geq\alpha)\right).
    \end{equation}

    Here, $\Vert$ and $R$ denote the operator of concatenation and the total number of adopted parcellated regions, respectively, while denoting $\mathbbm{1}(\cdot)$ as the indicator function that returns 1 if the given condition is true, or returns 0 otherwise. We define $\#$ as the count of elements in a given argument; thereby, $\#\mathcal{V}_r$ denotes the total number of voxel indices within $r$-th parcellated regions. With this, we obtain the significant ROI $\hat{\mathbf{x}}_q$ as AD-effect ROIs, which consist of $R$ number of ROIs as $\hat{\mathbf{x}}_q\in\mathbb{R}^{R\times1}$.

    \subsubsection{LiCoL architecture}
    We perform an objective classification task to verify the effectiveness of the acquired AD-effect ROIs $\hat{\mathbf{x}}_q$. Inspired by Transformer~\cite{NIPS2017_3f5ee243}, which could consider the global relationship, we devise a LiCoL to quantitatively account for the most influential ROIs, which contributed to the AD prediction among AD-effect ROIs (illustrated in Fig.~\ref{fig:licol}).

    In a nutshell, our proposed LiCoL maps a set of ROIs (\ie, the type of vector) comprised of query, key, and value onto the input. As AD-effect ROIs $\hat{\mathbf{x}}_q$ contain the characteristics of universal subregions where GM density differences between clinical stages are prominent concerning AD progression, we establish the query as AD-effect ROIs to adequately guide the attention within the LiCoL. Contrary to the query, the key and value are individually constructed for each training sample. Similar to AD-effect ROI acquisition in Eq.~(\ref{eq:ad-effect-roi}), the key and value would be defined as patient-wise ROIs $\hat{\mathbf{x}}_s\in\mathbb{R}^{R\times1}$, which have averaged the voxel values corresponding to the voxel indices $\mathcal{V}$ in each GM density map $\hat{\mathbf{X}}$ along the parcellated region. Subsequently, the query, key, and value are transformed into the set of embedded matrices $\mathbf{Q}, \mathbf{K}, \mathbf{V} \in \mathbb{R}^{R\times D}$ by multiplying ROIs with the respective embedding layer $\mathbf{w}\in\mathbb{R}^{1\times D}$. Specifically, the set of $\mathbf{Q}, \mathbf{K}, \mathbf{V}   $ represent the embedded matrices produced from a linear transformation, $\mathbf{Q}=\hat{\mathbf{x}}_q\mathbf{w}_q$, $\mathbf{K}=\hat{\mathbf{x}}_s\mathbf{w}_k$, and $\mathbf{V}=\hat{\mathbf{x}}_s\mathbf{w}_v$, respectively, where $\mathbf{w}_q, \mathbf{w}_k,$ and $\mathbf{w}_v$ are learnable embedding weights. We then compute the counterfactual-guided attention matrix $\mathbf{A}\in\mathbb{R}^{R\times D}$ as:
    \begin{equation}\label{eq:cf_guided_att}
        \mathbf{A}=\operatorname{Attention}(\mathbf{Q},~\mathbf{K},~\mathbf{V}) = g\left(\frac{\mathbf{Q}\mathbf{K}^\top}{\sqrt{d}}\right)\mathbf{V},
    \end{equation}
    where $\top$ and $d$ respectively denote a transpose operation, and the number of ROIs used from the key for scaling, while $g$ denotes the softmax function. Through Eq.~(\ref{eq:cf_guided_att}), our LiCoL allows for enforcing the predominant difference between clinical stages (\ie, inter-subject variability) while reflecting the discriminative attributes of the patient's individual (\ie, intra-subject variability).
    
    To infer the diagnostic prediction, we first apply the channel-wise mean pooling $\operatorname{MP}_{\rightarrow}$ to reshape the output of the counterfactual-guided attention map $\mathbf{A}$ to match the same size as the input query $\hat{\mathbf{x}}_q\in\mathbb{R}^{R\times 1}$, so that the size of the counterfactual-guided attention map pooled in the channel is defined as $\operatorname{MP}_{\rightarrow}(\mathbf{A})\in\mathbb{R}^{R\times 1}$. Thereafter, we employ a residual connection using an element-wise addition, followed by an ROI-wise mean pooling $\operatorname{MP}_{\downarrow}$ to obtain the final predicted label $\hat{\mathbf{y}}$ as the following:
    % Having applied the channel-wise mean pooling to reshape the output of the counterfactual-guided attention map to match the size of the input query $\mathbf{x}_q$, we then employ a residual connection using an element-wise addition, followed by an ROI-wise mean pooling to obtain the final predicted label $\bar{y}$ as:
    \begin{equation}\label{eq:LiCoL}
        \hat{\mathbf{y}} = \sigma\left(\operatorname{MP}_{\downarrow}\left(\operatorname{MP}_{\rightarrow}\left(\mathbf{A}\right)+\hat{\mathbf{x}}_q\right)\right)
    \end{equation}
    	where $\sigma$ denotes the softmax function. 
    	
    Eventually, we train the LiCoL by minimizing the classification loss via $\operatorname{CE}$ as:
    \begin{equation}
        \mathcal{L}_{\text{cls}}^{\text{LiCoL}} = \mathbb{E}_{\hat{\mathbf{X}}\sim P_{\hat{\mathbf{X}}}}\left[\operatorname{CE}\left(\hat{\mathbf{y}},\mathbf{y})\right)\right],\label{cls_loss}
    \end{equation}
    where $\hat{\mathbf{X}}$ and $\mathbf{y}$ denote a GM density map and its ground-truth label, respectively. It should be noted that during training, the LiCoL is learned employing both rGMs and cGMs, but it uses only test samples acquired from the rGMs $\hat{\mathbf{X}}^r$ to evaluate the performance of the LiCoL.

\section{Results and Analyses}\label{sec:results}
    \subsection{Datasets}
    The ADNI and GARD comprised 3D sMRIs and were acquired from various patient groups ranging from those with CN to those with MCI and AD. These three categories were annotated based on the standard clinical criteria, including mini-mental state examination (MMSE) scores and clinical dementia ratings. Although some subjects had accumulated multiple MRIs through their follow-up in both datasets, we only selected and utilized their baseline (first-visit) MRIs. The sMRI preprocessing and details of the subjects' demographic information are summarized in Supplementary A.
    
    The ADNI dataset consisted of 1,540 sMRI scans that included CN subjects (n = 433), MCI subjects (n = 748), and AD subjects (n = 359) from the combined ADNI-1 and ADNI-2 studies. As a special case for ADNI, two subgroups of MCI subjects were sampled: the progressive MCI (pMCI) (n = 251), including MCI subjects who converted to AD within 36 months of screening, and the stable MCI (sMCI) (n = 497), including those who remained in the MCI group within 36 months of screening. The baseline of the ADNI-1 included 1.5T T1-weighted sMRI scans acquired from 822 subjects, and ADNI-2 included 3T T1-weighted sMRI scans acquired from 718 subjects. Meanwhile, the baseline GARD dataset contained 1.5T T1-weighted sMRI scans acquired from 745 subjects, including CN subjects (n = 261), MCI subjects (n = 375), and AD subjects (n = 109).
    
    \subsection{Experimental setup}
    As the ADNI dataset consists of four categories (\ie, CN, sMCI, pMCI, and AD), we merged the subjects of the sMCI and pMCI as MCI to maintain consistency with the GARD dataset. \rev{Henceforth, we evaluated both datasets for four scenarios: (i) CN $vs.$ MCI, (ii) MCI $vs.$ AD, (iii) CN $vs.$ AD, and (iv) CN $vs.$ MCI $vs.$ AD. We further performed the CN $vs.$ sMCI $vs.$ pMCI $vs.$ AD scenario using the ADNI apart from the granularity issues of GARD and reported their results in Table S19 and Table S20.} All experiments and evaluations were conducted using five-fold cross-validation. 

    \subsubsection{Evaluation metrics}
    \rev{All experimental results for the predictive performance were quantitatively validated and evaluated using four criteria: the (multi-class) area under the receiver operating characteristic curve (mAUC or AUC), accuracy, sensitivity, and specificity.} Specifically, the best predictive model $\mathcal{C}$ for generating counterfactual images was selected by assessing the AUC during training, and the test performance corresponding to its model was reported in Supplementary B. Meanwhile, we further measured the predictive performance under these evaluation metrics to verify the effectiveness of our proposed LiCoL along with AD-effect ROIs. To affirm the generalizability of the cGM for augmentation, we used precision and recall, which were preferred over other alternatives when the class distribution was significantly skewed~\cite{davis2006relationship}. 
    % \ks{Compared with each baseline, we further performed a Wilcoxon signed-rank test~\cite{woolson2007wilcoxon} with $p$-values of 0.05 and 0.01 for both AUC and accuracy to demonstrate the statistical significance of our proposed method.}

    \subsubsection{Statistical hypothesis test}
    For the AD-effect map acquisition and augmentation of cGMs for the statistical map, we measured the statistical significance of categorizing with the same population using a two-sample \textit{t}-test ($p$-value $<$ 0.05) to group the same clinical stage of the rGMs and cGMs. This statistical test indicates that two independent samples have identical expected values. Thus, it can be assumed that the cGMs are adequately mapped on the identical distribution of rGMs, respectively. In a nutshell, the AD-labeled cGM and the AD-labeled rGM were considered the same AD category.

    \begin{figure}[t]
        \centering
        \subfloat[Comparison of the generated CF maps with ground-truth maps. In the first column for each scenario, the number at the top left of the respective image indicates the image ID. We also overlay purple, green, and orange boxes to highlight the ventricular, cortex, and hippocampal regions, which are closely related to the progression of MCI and AD.]{\label{fig:cf-map-result}{\includegraphics[width=.46\textwidth ]{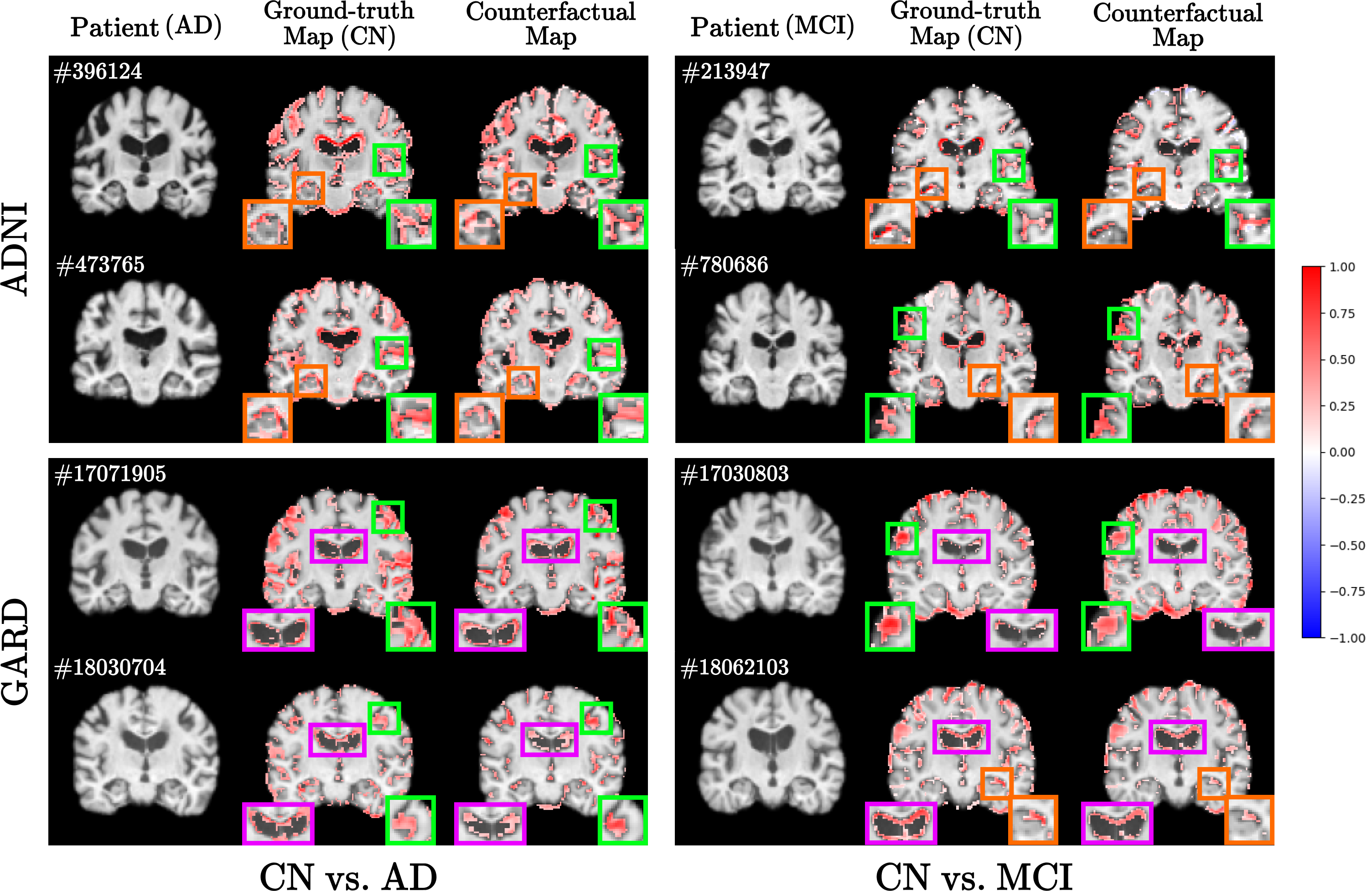}}}
        \vspace{0.1cm}
        \subfloat[Results of normalized cross-correlation (NCC) scores. We differentiated the NCC scores for the respective direction scenario of the generated CF map. Here, we defined the ground truth and the CF maps for various conversion scenarios, including CN←MCI, MCI←AD, and CN←AD as ``+'', and CN→MCI, MCI→AD, and CN→AD as ``-'', and we calculated NCC(+) and NCC(-) for each, respectively.]{\label{fig:ncc-score-result}{\includegraphics[width=.46\textwidth ]{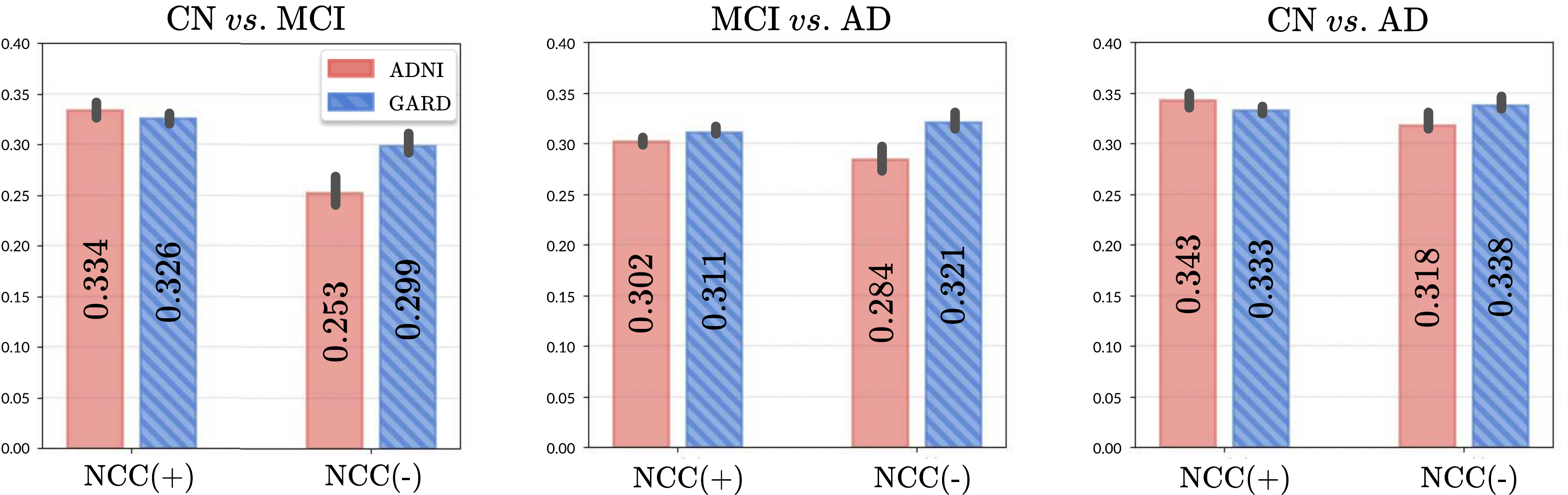}}}
        \caption{Qualitative and quantitative evaluations of counterfactual reasoning on ADNI and GARD datasets.}
    \label{fig:cf-reasoning-result}
    \end{figure}

    \subsection{Verification of counterfactual images}
    We conducted qualitative and quantitative assessments to validate the reliability of counterfactual reasoning on the ADNI and GARD datasets. As no ground truth is available to evaluate the generated CF map, we constructed the ``\emph{pseudo}'' ground-truth map (\ie, an observed disease progression map), defined as the difference between longitudinal samples of two different clinical stages over the same patient (refer to Supplementary D). For example, if the CN subject is selected as a baseline, either the MCI or AD subject shall be defined as a counterpart of the target image, depending on the scenario. This ground-truth map (second column in each scenario of Fig.~\ref{fig:cf-map-result}) exhibited an excellent representation of atrophies, indicating which regions were changed according to the clinical stage conversion.
    
    In Fig.~\ref{fig:cf-map-result}, we illustrate the CF map examples generated by our CMG~\cite{oh2022learn} from the tasks of CN $vs.$ AD and CN $vs.$ MCI classification. The generated CF maps on both the ADNI and GARD longitudinal samples showed excellent agreement with their corresponding pseudo-ground-truth map. In fact, the hippocampal region (orange box) is known to be the most prominent MCI/AD region~\cite{zhao2020independent,risacher2009baseline}; however, it is also known that morphological variations due to MCI/AD in many other regions are also involved. Regarding these disease characteristics in the brain, we observed that our CMG successfully captured reduced ventricle and hypertrophy in the hippocampus~\cite{jack2004comparison} while containing subtle variations that occurred in most cerebral cortex regions.
    
    As a numerical assessment metric of the generated CF maps, we calculated the NCC score in Fig.~\ref{fig:ncc-score-result} to measure the morphological and structural similarity between a generated CF map and a pseudo-ground-truth map as proposed by~\cite{oh2022learn}. We leveraged the NCC score because it has the primary property of not being sensitive to the magnitude of the signals. Compared with the LEAR's performance~\cite{oh2022learn}, which reported the ADNI performance only, the NCC scores of GARD were comparable to those on the ADNI in all scenarios. Based on these NCC scores, we hypothesized that the generated CF maps on both ADNI and GARD datasets captured pathologically subtle changes associated with AD progression, yielding the CMG's scalability over two independent and different racial data sources. As an extensive analysis, we further experimented with the cross-domain scheme across the ADNI and GARD in Supplementary B and compared the accuracy of the CMG when precisely predicting the synthesized images as the target label in Supplementary I. In addition, we have thoroughly conducted ablation studies of an entire pipeline according to the losses used in the CMG training to quantitatively and qualitatively explore how the quality of synthesized images influences classification performance in Supplementary J. Leveraged by those qualitatively and quantitatively promising results, we can justify our proposed framework to exploit the counterfactually-generated samples and accept our findings from the experimental analysis described below.
    
    % \clearpage
    \subsection{AD-effect map and ROI feature representation}
\begin{figure*}[t]
    \centering
    \includegraphics[width=.95\textwidth]{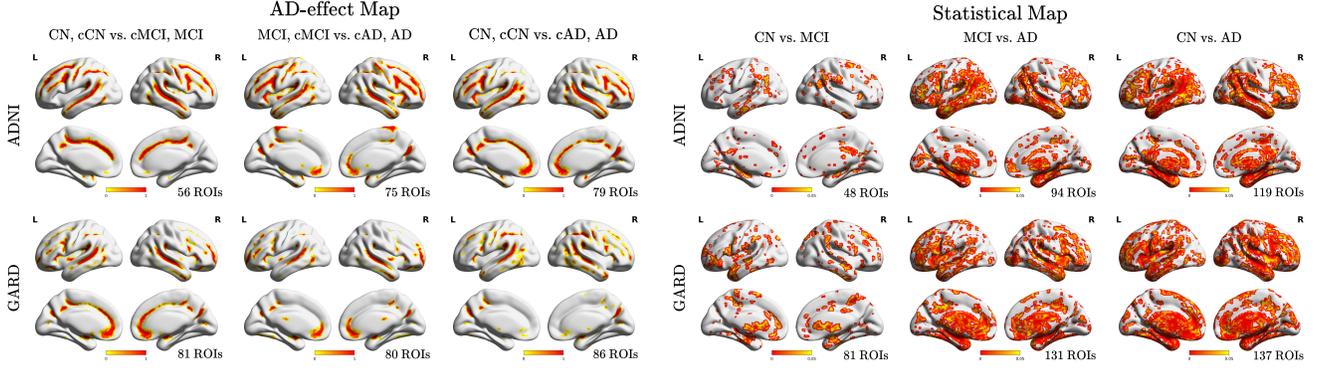}
    \caption{Illustration of inferred AD-effect and statistical maps on ADNI and GARD datasets in binary scenarios. The cCN, cMCI, and cAD are abbreviations of the CN-/MCI-/AD-labeled cGM, respectively. The color scale in AD-effect maps (left) represents normalized GM density, while the color scale in statistical maps (right) represents \textit{p}-values from a two-sample \textit{t}-test. We also present the respective numbers of significant ROIs derived from each AD-effect and statistical map below.}
    \label{fig:ad_cau_map}
    \end{figure*}

\begin{figure}[t]
        \centering
        \includegraphics[width=.5\textwidth]{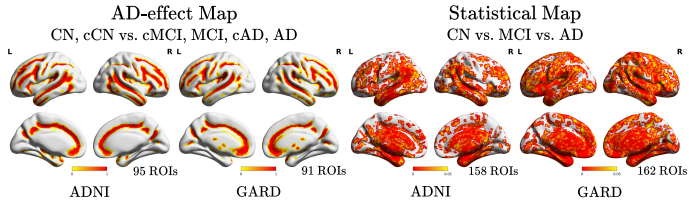}
        \caption{Illustration of inferred AD-effect and statistical maps on ADNI and GARD datasets in the multi-class scenario. The color scale in AD-effect maps (left) represents normalized GM density, while the color scale in statistical maps (right) represents $p$-values from the ANOVA.}
        \label{fig:ad_cau_map_multi}
    \end{figure}

    \rev{Using Eq.~(\ref{eq:AD-effect-map}), we estimated AD-effect maps for binary scenarios that identify cortical and subcortical brain regions, where the GM density contrasts were considerably differentiated between clinical stages (results of AD-effect maps in Fig.~\ref{fig:ad_cau_map}). Similar to the ensemble strategy, we gathered the rGMs and cGMs used in each binary scenario to generate AD-effect maps over the multi-class scenario. Subsequently, we partitioned them into normal (\ie, CN, cCN) and patient (\ie, cMCI, MCI, cAD, AD) groups and calculated the AD-effect map via Eq.~(\ref{eq:AD-effect-map}) in the same way as in binary scenarios (see Fig.~\ref{fig:ad_cau_map_multi}). By doing so, this map is possible to fully cover the disease progression across the AD spectrum. In particular, the merit of this strategy is to incorporate the outcomes of each binary scenario, which can focus on distinguishing between specific pairs of classes, allowing it to specialize in maximizing the discriminative power between those classes, potentially improving accuracy for distinguishing specific disease heterogeneity. For comparison, we computed a conventional group-level statistical map ($p$-value $<$ 0.01) by measuring the two-sample \textit{t}-test between different stages of rGMs for binary scenarios and ANOVA among all disease stages of rGMs for multi-class scenarios (results of statistical maps in Fig.~\ref{fig:ad_cau_map} and Fig.~\ref{fig:ad_cau_map_multi}). Note that the AD-effect map and statistical map used a 0.99 percentile threshold and $p$-value $<$ 0.01, respectively, which showed the highest validation performance across all scenarios (see Table S13).}
    % \ks{Using Eq.~(\ref{eq:AD-effect-map}), we first estimated AD-effect maps for binary scenarios that identify cortical and subcortical brain regions, where the GM density contrasts were considerably differentiated between clinical stages (results of AD-effect maps in Fig.~\ref{fig:ad_cau_map}). For the multi-class scenario, we gathered the rGMs and cGMs used in each binary scenario, similar to an ensemble strategy, and partitioned them into normal (\ie, CN, cCN) and patient (\ie, cMCI, MCI, cAD, AD) groups. Subsequently, the AD-effect map for the multi-class scenario was obtained using Eq.~(\ref{eq:AD-effect-map}) in the same way as in binary scenarios (see Fig.~\ref{fig:ad_cau_map_multi}). By doing so, it can fully cover the disease progression across the AD spectrum. For comparison, we computed a conventional group-level statistical map ($p$-value $<$ 0.01) by measuring the two-sample \textit{t}-test between different stages of rGMs for binary scenarios and ANOVA among all disease stages for a multi-class scenario (results of statistical maps in Fig.~\ref{fig:ad_cau_map} and Fig.~\ref{fig:ad_cau_map_multi}). Note that the AD-effect map and statistical map used a 0.99 percentile threshold and $p$-value $<$ 0.01, respectively, which showed the highest validation performance across all scenarios (see Table S13).}

    Regarding feature engineering, we considered these AD-effect and statistical maps to identify the class-discriminative areas and applied different thresholding strategies (\ie, percentile $vs.$ $p$-value) for ROI selection. For the configuration of statistical ROIs $\mathcal{R}_{\text{stat}}$, we used Eq.~(\ref{eq:ad-effect-roi}), substituted from the AD-effect map to the statistical map. Although such AD-effect and statistical ROIs are produced at the equivalent criterion, the fundamental disparity in acquiring significant ROIs is provoked by each property underlying the AD-effect map and statistical map. The AD-effect map aims to discern regions of the brain with \textit{atrophy} and \textit{hypertrophy} that are the most drastically activated, whereas the statistical map takes into account the \textit{probability} of observing a difference in disease progression.
    
    \ks{In Fig.~\ref{fig:ad_cau_map} and Fig.~\ref{fig:ad_cau_map_multi}, the AD-effect map showed a more fine-grained regional localization than the statistical map. Whereas some areas were identically prominent in the AD-effect and statistical maps, other regions were revealed by only one map. Based on the classification performance in Table S13, we assume that the AD-effect map has better-localized regions than the statistical map. On the one hand, the statistically significant areas do not necessarily reflect class-discriminative information at an individual-level prediction. On the other hand, our proposed AD-effect map finds class-discriminative areas, which may not be statistically significant in a size-limited dataset by distilling the augmented cGMs. Intriguingly, according to the scenario in which the disease progressed from CN to AD, the number of AD-effect ROI and statistical ROI showed a tendency to increase gradually. One possible explanation for such differences is the clinical heterogeneity between MCI and AD based on disease severity. This could be supported by our observation that the MCI group in both ADNI and GARD datasets had fewer significant ROIs in the CN $vs.$ MCI scenario, despite having a larger amount of samples than the AD group (see Table S1). Therefore, we hypothesize that this scenario would have resulted in fewer ROIs because of the difficulty in highlighting the dementia-manifested regions within CN and MCI owing to anatomically subtle variations. The effectiveness and thorough analysis of these significant ROIs are described in the following subsections.}
    % \rev{Intriguingly, AD-effect ROIs and statistical ROIs gradually increased according to the scenario in which the disease proceeded from CN to AD. We hypothesize that differences may be attributed to the clinical heterogeneity based on the disease severity between MCI and AD. Indeed, the MCI group on both ADNI and GARD datasets has smaller significant ROIs in the CN $vs.$ MCI scenario, despite having a larger amount of samples than the AD group (see Table S1). From this observation, we assume that the CN $vs.$ MCI scenario would have constituted fewer ROIs because of the difficulty in highlighting the dementia-manifested regions within CN and MCI owing to anatomically subtle variations. The efficiency and thorough analysis of these significant ROIs are described in the following subsections.}

\subsection{Predictive performance evaluation}\label{sec: score eval}
        We performed exhaustive experiments on the predictive performance of our proposed LiCoL and the effectiveness of the AD-effect ROI set $\mathcal{R}_{\text{eff}}$. As competing methods, we have adopted ML- and DL-based models, which possessed inherent interpretability for the model's decision or achieved state-of-the-art AD classification performance. Refer to Supplementary C for the implementation details and hyperparameters for baselines and our LiCoL training. 

    \begin{table*}[t]\footnotesize \setlength{\tabcolsep}{2.5pt}
        \centering
        \caption{The classification results compared with the ML-/DL-based models in various binary scenarios on ADNI and GARD datasets with four evaluation criteria (\ie, AUC, accuracy, sensitivity, and specificity). Here, these baselines were trained using both rGM and cGM for a fair comparison. The symbols * and ** for AUC and accuracy denote the statistical significance via the Wilcoxon signed-rank test at $p$ $<$ 0.05 and $p$ $<$ 0.01, respectively, when comparing our model’s performance with each baseline. The highest scores are in boldface, and the second-highest scores are underlined.}
        \label{tab:density_performance_with_cGM}
        \scalebox{0.95}{
        \begin{tabular}{ccccccc|cccc}
        \toprule
        & & & \multicolumn{4}{c}{{\textbf{ADNI}}} & \multicolumn{4}{c}{{\textbf{GARD}}}\\
         & \multicolumn{1}{c}{\textbf{Model}} & \multicolumn{1}{c}{\textbf{Params}} & \multicolumn{1}{c}{\textbf{AUC}} & \multicolumn{1}{c}{\textbf{Accuracy}} & \multicolumn{1}{c}{\textbf{Sensitivity}} & \multicolumn{1}{c}{\textbf{Specificity}} & \multicolumn{1}{c}{\textbf{AUC}} & \multicolumn{1}{c}{\textbf{Accuracy}} & \multicolumn{1}{c}{\textbf{Sensitivity}} & \multicolumn{1}{c}{\textbf{Specificity}}\\
        \cmidrule(lr){2-2} \cmidrule(lr){3-3} \cmidrule(lr){4-7} \cmidrule(lr){8-11}
        \parbox[t]{3mm}{\multirow{7}{*}{\rotatebox[origin=c]{90}{CN $vs.$ MCI}}} & \parbox[t]{6mm}{\multirow{1}{*}{\rotatebox[origin=c]{0}{SVM}}} & \text{-} & \enspace\enspace\text{0.6721}$\pm$\text{0.02}$^{**}$ &\enspace\enspace\text{0.6882}$\pm$\text{0.03}$^{**}$ & \text{0.6655}$\pm$\text{0.03} & \text{0.7385}$\pm$\text{0.01} & \enspace\enspace\text{0.7145}$\pm$\text{0.02}$^{**}$ &\enspace\enspace\text{0.7059}$\pm$\text{0.03}$^{**}$ & \text{0.7281}$\pm$\text{0.05} & \text{0.6798}$\pm$\text{0.03}\\
        & \parbox[t]{12mm}{\multirow{1}{*}{ResNet18}} & \text{33.17M} & \enspace\text{0.7178}$\pm$\text{0.03}$^{*}$ &\enspace\text{0.7164}$\pm$\text{0.02}$^{*}$ & \text{0.7215}$\pm$\text{0.05} & \text{0.6976}$\pm$\text{0.04} & \enspace\text{0.7395}$\pm$\text{0.03}$^{*}$ &\enspace\text{0.7211}$\pm$\text{0.03}$^{*}$ & \text{0.7385}$\pm$\text{0.05} & \text{0.7661}$\pm$\text{0.04}\\
        & \parbox[t]{13.5mm}{\multirow{1}{*}{ResAttNet}} & \text{64.13M} & \text{0.7345}$\pm$\text{0.03} &\text{0.7259}$\pm$\text{0.03} & \textbf{0.7489}$\pm$\textbf{0.02} & \text{0.7398}$\pm$\text{0.02} & \text{0.7658}$\pm$\text{0.02} &\text{0.7621}$\pm$\text{0.03} & \text{0.7285}$\pm$\text{0.04} & \text{0.8013}$\pm$\text{0.02}\\
        & \parbox[t]{5mm}{\multirow{1}{*}{ViT}} & 33.87M & \enspace\enspace\text{0.7071}$\pm$\text{0.06}$^{**}$ &\enspace\text{0.7119}$\pm$\text{0.06}$^{*}$ & \text{0.6985}$\pm$\text{0.04} & \text{0.7481}$\pm$\text{0.05} & \enspace\enspace\text{0.7153}$\pm$\text{0.04}$^{**}$ &\enspace\text{0.7218}$\pm$\text{0.04}$^{*}$ & \text{0.6987}$\pm$\text{0.05} & \text{0.7436}$\pm$\text{0.04}\\
        & \parbox[t]{6mm}{\multirow{1}{*}{M3T}} & 29.12M & \text{0.7453}$\pm$\text{0.03} &\text{0.7389}$\pm$\text{0.03} & \underline{0.7274$\pm$0.05} & \text{0.7748}$\pm$\text{0.04} & \text{0.7698}$\pm$\text{0.03} &\text{0.7657}$\pm$\text{0.03} & \text{0.7315}$\pm$\text{0.04} & \textbf{0.8304}$\pm$\textbf{0.04}\\
        & \parbox[t]{12.5mm}{\multirow{1}{*}{DSTANet}} & \text{3.04M} & \underline{0.7592$\pm$0.03} &\underline{0.7448$\pm$0.02} & \text{0.7093}$\pm$\text{0.05} & \underline{0.8142$\pm$0.04} & \underline{0.7758$\pm$0.04} &\textbf{0.7690}$\pm$\textbf{0.03} & \underline{0.7408$\pm$0.05} & \text{0.8189}$\pm$\text{0.04}\\
        % \cmidrule(lr){2-11}
        & \parbox[t]{8mm}{\multirow{1}{*}{\rotatebox[origin=c]{0}{\textbf{LiCoL}}}} & \text{1,536} & \textbf{0.7678}$\pm$\textbf{0.01} &\textbf{0.7562}$\pm$\textbf{0.02} & \text{0.7143}$\pm$\text{0.02} & \textbf{0.8182}$\pm$\textbf{0.02} & \textbf{0.7778}$\pm$\textbf{0.02} &\underline{0.7662$\pm$0.02} & \textbf{0.7485}$\pm$\textbf{0.03} & \underline{0.8246$\pm$0.02}\\
        \cmidrule(lr){1-11}
        \parbox[t]{3mm}{\multirow{7}{*}{\rotatebox[origin=c]{90}{MCI $vs.$ AD}}} & \parbox[t]{6mm}{\multirow{1}{*}{\rotatebox[origin=c]{0}{SVM}}} & \text{-} & \enspace\text{0.7519}$\pm$\text{0.01}$^{*}$  &\enspace\text{0.7329}$\pm$\text{0.02}$^{*}$ & \text{0.6657}$\pm$\text{0.03} & \text{0.7785}$\pm$\text{0.03} & \enspace\text{0.7284}$\pm$\text{0.03}$^{*}$ &\enspace\text{0.7386}$\pm$\text{0.03}$^{*}$ & \text{0.8329}$\pm$\text{0.04} & \text{0.7391}$\pm$\text{0.04}\\
        & \parbox[t]{12mm}{\multirow{1}{*}{ResNet18}} & \text{33.17M} & \text{0.7705}$\pm$\text{0.01} &\enspace\text{0.7551}$\pm$\text{0.04}$^{*}$ & \text{0.6971}$\pm$\text{0.06} & \text{0.8495}$\pm$\text{0.05} & \enspace\text{0.7569}$\pm$\text{0.03}$^{*}$ &\text{0.7729}$\pm$\text{0.04} & \text{0.6619}$\pm$\text{0.04} & \textbf{0.8183}$\pm$\textbf{0.05}\\
        & \parbox[t]{13.5mm}{\multirow{1}{*}{ResAttNet}} & \text{64.13M} & \text{0.7719}$\pm$\text{0.03} &\text{0.7895}$\pm$\text{0.02} & \text{0.7814}$\pm$\text{0.03} & \text{0.8824}$\pm$\text{0.03} & \text{0.7719}$\pm$\text{0.03} &\textbf{0.7932}$\pm$\textbf{0.02} & \text{0.6918}$\pm$\text{0.06} & \text{0.8096}$\pm$\text{0.03}\\
        & \parbox[t]{5mm}{\multirow{1}{*}{ViT}} & 33.87M & \enspace\text{0.7458}$\pm$\text{0.06}$^{*}$ &\enspace\enspace\text{0.7179}$\pm$\text{0.05}$^{**}$ & \text{0.6321}$\pm$\text{0.11} & \text{0.8195}$\pm$\text{0.13} & \enspace\text{0.7392}$\pm$\text{0.06}$^{*}$ &\enspace\text{0.7328}$\pm$\text{0.04}$^{*}$ & \text{0.7089}$\pm$\text{0.07} & \underline{0.8134$\pm$0.08}\\
        & \parbox[t]{6mm}{\multirow{1}{*}{M3T}} & 29.12M & \textbf{0.7788}$\pm$\textbf{0.03} &\underline{0.7903$\pm$0.04} & \underline{0.7989$\pm$0.05} & \text{0.8887}$\pm$\text{0.04} & \textbf{0.7811}$\pm$\textbf{0.04} &\text{0.7795}$\pm$\text{0.03} & \text{0.7395}$\pm$\text{0.05} & \text{0.7942}$\pm$\text{0.04}\\
        & \parbox[t]{12.5mm}{\multirow{1}{*}{DSTANet}} & \text{3.04M} & \text{0.7686}$\pm$\text{0.03} &\text{0.7869}$\pm$\text{0.04} & \text{0.7759}$\pm$\text{0.05} & \textbf{0.8981}$\pm$\textbf{0.03} & \text{0.7698}$\pm$\text{0.04} &\text{0.7771}$\pm$\text{0.03} & \underline{0.8659$\pm$0.04} & \text{0.7927}$\pm$\text{0.03}\\
        % \cmidrule(lr){2-11}
        & \parbox[t]{8mm}{\multirow{1}{*}{\rotatebox[origin=c]{0}{\textbf{LiCoL}}}} & \text{1,536} & \underline{0.7727$\pm$0.02} &\textbf{0.7921}$\pm$\textbf{0.02} & \textbf{0.8014}$\pm$\textbf{0.02} & \underline{0.8913$\pm$0.02} & \underline{0.7753$\pm$0.02} &\underline{0.7814$\pm$0.02} & \textbf{0.8917}$\pm$\textbf{0.01} & \text{0.7955}$\pm$\text{0.02}\\
        \cmidrule(lr){1-11}
        \parbox[t]{3mm}{\multirow{7}{*}{\rotatebox[origin=c]{90}{CN $vs.$ AD}}} & \parbox[t]{6mm}{\multirow{1}{*}{\rotatebox[origin=c]{0}{SVM}}} & \text{-} & \text{0.9167}$\pm$\text{0.01} &\text{0.9167}$\pm$\text{0.01} & \text{0.8889}$\pm$\text{0.02} & \text{0.9444}$\pm$\text{0.02} & \enspace\enspace\text{0.8774}$\pm$\text{0.02}$^{**}$ &\enspace\enspace\text{0.8533}$\pm$\text{0.03}$^{**}$ & \text{0.8912}$\pm$\text{0.02} & \text{0.8778}$\pm$\text{0.01}\\
        & \parbox[t]{12mm}{\multirow{1}{*}{ResNet18}} & \text{33.17M} & \text{0.9265}$\pm$\text{0.03} &\text{0.9219}$\pm$\text{0.04} & \text{0.9078}$\pm$\text{0.03} & \text{0.9209}$\pm$\text{0.05} & \enspace\text{0.8930}$\pm$\text{0.03}$^{*}$ &\text{0.8913}$\pm$\text{0.03} & \text{0.8733}$\pm$\text{0.05} & \text{0.9093}$\pm$\text{0.04}\\
        & \parbox[t]{13.5mm}{\multirow{1}{*}{ResAttNet}} & \text{64.13M} & \text{0.9267}$\pm$\text{0.01} &\text{0.9186}$\pm$\text{0.02} & \text{0.8889}$\pm$\text{0.03} & \text{0.9544}$\pm$\text{0.03} & \text{0.9254}$\pm$\text{0.02} &\text{0.9117}$\pm$\text{0.02} & \text{0.9168}$\pm$\text{0.02} & \text{0.9169}$\pm$\text{0.03}\\
        & \parbox[t]{5mm}{\multirow{1}{*}{ViT}} & 33.87M & \text{0.9187}$\pm$\text{0.07} &\text{0.9183}$\pm$\text{0.04} & \text{0.8970}$\pm$\text{0.05} & \text{0.9415}$\pm$\text{0.06} & \text{0.9052}$\pm$\text{0.04} &\text{0.9103}$\pm$\text{0.03} & \underline{0.9314$\pm$0.04} & \text{0.8921}$\pm$\text{0.05}\\
        & \parbox[t]{6mm}{\multirow{1}{*}{M3T}} & 29.12M & \underline{0.9344$\pm$0.03} &\textbf{0.9301}$\pm$\textbf{0.04} & \textbf{0.9311}$\pm$\textbf{0.03} & \underline{0.9569$\pm$0.03} & \underline{0.9279$\pm$0.03} &\textbf{0.9215}$\pm$\textbf{0.02} & \text{0.9032}$\pm$\text{0.04} & \textbf{0.9433}$\pm$\textbf{0.02}\\
        & \parbox[t]{12.5mm}{\multirow{1}{*}{DSTANet}} & \text{3.04M} & \text{0.9286}$\pm$\text{0.03} &\text{0.9237}$\pm$\text{0.03} & \text{0.9185}$\pm$\text{0.04} & \text{0.9533}$\pm$\text{0.04} & \text{0.9271}$\pm$\text{0.04} &\text{0.9137}$\pm$\text{0.02} & \text{0.9287}$\pm$\text{0.04} & \text{0.8968}$\pm$\text{0.06}\\
        % \cmidrule(lr){2-11}
        & \parbox[t]{8mm}{\multirow{1}{*}{\rotatebox[origin=c]{0}{\textbf{LiCoL}}}} & \text{1,536} & \textbf{0.9371}$\pm$\textbf{0.01} &\underline{0.9285$\pm$0.01} & \underline{0.9271$\pm$0.02} & \textbf{0.9591}$\pm$\textbf{0.02} & \textbf{0.9304}$\pm$\textbf{0.01} &\underline{0.9193$\pm$0.01} & \textbf{0.9441}$\pm$\textbf{0.01} & \underline{0.9283$\pm$0.01}\\
        \bottomrule
        \end{tabular}}
    \end{table*}

\subsubsection{Quantitative comparison with ML methods}
        \rev{We employed six ML models, \ie, support vector machine (SVM), random forest, logistic regression, decision tree, Lasso regression, and Ridge regression. Table~\ref{tab:density_performance_with_cGM} and Table~\ref{tab: multi-cls} present the results of SVM, as it achieves the best performance among the comparative ML baselines\footnote{The other ML baselines' performance using $\mathcal{R}_{\text{eff}}$ and $\mathcal{R}_{\text{stat}}$ is reported in Table S14, Table S15, and Table S17, respectively.}. Note that the SVM performance reported in these tables represents the results trained with $\mathcal{R}_{\text{eff}}$ for a fair comparison with our LiCoL.
        Nonetheless, our method outperformed the mean AUC (ADNI: 4.56\%$\uparrow$ and GARD: 5.44\%$\uparrow$) and mean accuracy (ADNI: 4.63\%$\uparrow$ and GARD: 5.64\%$\uparrow$) of SVM in all binary scenarios with statistical significance. In addition, even in the multi-class scenario, our LiCoL has still shown outstanding performance in mAUC (ADNI: 4.62\%$\uparrow$ and GARD: 4.75\%$\uparrow$) and accuracy (ADNI: 3.40\%$\uparrow$ and GARD: 3.39\%$\uparrow$) over SVM. Most remarkably, utilizing $\mathcal{R}_{\text{eff}}$ achieved better performance improvement in the CN $vs.$ MCI and the CN $vs.$ MCI $vs.$ AD scenario compared to $\mathcal{R}_{\text{stat}}$ performance (refer to Table S14, Table S15, and Table S17), despite being difficult to identify disease-affected regions associated with brain atrophies owing to subtle anatomical variations. Having observed these predictive outcomes, we confirmed that our discovered $\mathcal{R}_{\text{eff}}$ played a critical role in revealing a relatively substantial difference in GM density. Thus, this piece of evidence assisted in exploiting $\mathcal{R}_{\text{eff}}$ because of its promise to distinguish most class-discriminative regions.}

    \begin{table}[t]\footnotesize \setlength{\tabcolsep}{1.5pt}
    \caption{The results of mAUC and accuracy on multi-class classification compared with the ML-/DL-based models on ADNI and GARD. When comparing LiCoL performance with each baseline, the symbols * and ** for mAUC and accuracy denote the statistical significance via the Wilcoxon signed-rank test at $p$ $<$ 0.05 and $p$ $<$ 0.01, respectively.}
    \centering
    \scalebox{0.87}{
    \begin{tabular}{cccc|cc}
    \toprule
    & & \multicolumn{2}{c}{\textbf{ADNI}} & \multicolumn{2}{c}{\textbf{GARD}}\\
    \parbox[t]{3mm}{\multirow{6}{*}{\rotatebox[origin=c]{90}{CN $vs.$ MCI $vs.$ AD}}} & \multicolumn{1}{c}{\textbf{Model}} & \multicolumn{1}{c}{\textbf{mAUC}} & \multicolumn{1}{c}{\textbf{Accuracy}} & \multicolumn{1}{c}{\textbf{mAUC}} & \multicolumn{1}{c}{\textbf{Accuracy}} \\
    \cmidrule(lr){2-2} \cmidrule(lr){3-4} \cmidrule(lr){5-6}
     & SVM &\enspace\text{0.7357}$\pm$\text{0.03}$^*$ &\enspace\text{0.6106}$\pm$\text{0.05}$^*$ &\text{0.7532}$\pm$\text{0.03} &\text{0.6235}$\pm$\text{0.05}\\
    & ResNet18 &\enspace\enspace\text{0.7153}$\pm$\text{0.05}$^{**}$ &\enspace\enspace\text{0.5729}$\pm$\text{0.08}$^{**}$ &\enspace\text{0.7248}$\pm$\text{0.06}$^{*}$ &\enspace\enspace\text{0.5823}$\pm$\text{0.07}$^{**}$\\
    & ResAttNet &\enspace\enspace\text{0.7232}$\pm$\text{0.04}$^{**}$ &\enspace\text{0.6096}$\pm$\text{0.05}$^{*}$ &\enspace\text{0.7313}$\pm$\text{0.05}$^*$ &\enspace\text{0.6037}$\pm$\text{0.06}$^*$\\
    & ViT &\enspace\text{0.7258}$\pm$\text{0.04}$^*$ &\enspace\text{0.6017}$\pm$\text{0.03}$^*$ &\enspace\text{0.7278}$\pm$\text{0.04}$^*$ &\enspace\text{0.5973}$\pm$\text{0.04}$^*$\\
    & M3T &\text{0.7548}$\pm$\text{0.04} &\text{0.6286}$\pm$\text{0.04} &\text{0.7612}$\pm$\text{0.04} &\text{0.6317}$\pm$\text{0.05}\\
    & DSTANet &\text{0.7784}$\pm$\text{0.04} &\text{0.6488}$\pm$\text{0.03} &\text{0.7859}$\pm$\text{0.03}&\text{0.6538}$\pm$\text{0.05}\\
    & \textbf{LiCoL} &\textbf{0.7919}$\pm$\textbf{0.03} &\textbf{0.6681}$\pm$\textbf{0.04} &\textbf{0.7972}$\pm$\textbf{0.04} &\textbf{0.6674}$\pm$\textbf{0.04}\\
    \bottomrule
    \end{tabular}}
    \label{tab: multi-cls}
    \end{table}

    \begin{table}[t]\footnotesize \setlength{\tabcolsep}{4pt}
    \centering
    \caption{Comparison of \rev{(m)AUC performance} between the AD-effect ROIs $\mathcal{R}_{\text{eff}}$ (Ours) and statistical ROIs $\mathcal{R}_{\text{stat}}$ \rev{when exploiting the common or additional discrepant ROIs. For the evaluation, we exploited our LiCoL as a classifier.}}
    \label{tab:invest_discrepant_rois}
    \scalebox{.84}{\begin{tabular}{cccccc}
    \toprule
    \multicolumn{1}{c}{\multirow{2}{*}{\textbf{Scenarios}}} & \multicolumn{1}{c}{\multirow{2}{*}{\textbf{Data}}} & \multicolumn{1}{c}{\multirow{2}{*}{\textbf{ROIs}}} & \multicolumn{3}{c}{\textbf{Categories}}\\
    \cmidrule(lr){4-6}
    & & & \textbf{Baseline} & \textbf{Intersection} & \textbf{Union}\\
    \midrule
    {\multirow{4.2}{*}{\rotatebox[origin=c]{0}{CN $vs.$ MCI}}} & {\multirow{2}{*}{\rotatebox[origin=c]{0}{ADNI}}} & \textbf{$\mathcal{R}_{\text{stat}}$} & \text{0.7222}$\pm$\text{0.02} & \text{0.7284}$\pm$\text{0.01} & \textbf{0.7389}$\pm$\text{0.03}\\
    & & \textbf{Ours} & \textbf{0.7678}$\pm$\text{0.01} & \text{0.7482}$\pm$\text{0.03} & \text{0.7604}$\pm$\text{0.02}\\
    \cmidrule(lr){2-6}
    & {\multirow{2}{*}{\rotatebox[origin=c]{0}{GARD}}} & \textbf{$\mathcal{R}_{\text{stat}}$} & \text{0.7385}$\pm$\text{0.02} & \text{0.7456}$\pm$\text{0.02} & \textbf{0.7479}$\pm$\text{0.02}\\
    & & \textbf{Ours} & \textbf{0.7778}$\pm$\text{0.02} & \text{0.7592}$\pm$\text{0.02} & \text{0.7745}$\pm$\text{0.03}\\
    \midrule
    {\multirow{4.2}{*}{\rotatebox[origin=c]{0}{MCI $vs.$ AD}}} & {\multirow{2}{*}{\rotatebox[origin=c]{0}{ADNI}}} & \textbf{$\mathcal{R}_{\text{stat}}$} & \text{0.7514}$\pm$\text{0.02} & \text{0.7538}$\pm$\text{0.02} & \textbf{0.7685}$\pm$\text{0.03}\\
    & & \textbf{Ours} & \textbf{0.7727}$\pm$\text{0.02} & \text{0.7727}$\pm$\text{0.02} & \text{0.7694}$\pm$\text{0.02}\\
    \cmidrule(lr){2-6}
    & {\multirow{2}{*}{\rotatebox[origin=c]{0}{GARD}}} & \textbf{$\mathcal{R}_{\text{stat}}$} & \text{0.7497}$\pm$\text{0.04} & \text{0.7512}$\pm$\text{0.03} & \textbf{0.7581}$\pm$\text{0.03}\\
    & & \textbf{Ours} & \textbf{0.7753}$\pm$\text{0.02} & \text{0.7693}$\pm$\text{0.01} & \text{0.7737}$\pm$\text{0.02}\\
    \midrule
    {\multirow{4.2}{*}{\rotatebox[origin=c]{0}{CN $vs.$ AD}}} & {\multirow{2}{*}{\rotatebox[origin=c]{0}{ADNI}}} & \textbf{$\mathcal{R}_{\text{stat}}$} & \text{0.8912}$\pm$\text{0.02} & \text{0.8933}$\pm$\text{0.01} & \textbf{0.9146}$\pm$\text{0.01}\\
    & & \textbf{Ours} & \textbf{0.9371}$\pm$\text{0.01} & \text{0.9371}$\pm$\text{0.01} & \text{0.9308}$\pm$\text{0.02}\\
    \cmidrule(lr){2-6}
    & {\multirow{2}{*}{\rotatebox[origin=c]{0}{GARD}}} & \textbf{$\mathcal{R}_{\text{stat}}$} & \text{0.9013}$\pm$\text{0.01} & \text{0.9074}$\pm$\text{0.01} & \textbf{0.9211}$\pm$\text{0.01}\\
    & & \textbf{Ours} & \textbf{0.9304}$\pm$\text{0.01} & \text{0.9304}$\pm$\text{0.01} & \text{0.9277}$\pm$\text{0.02}\\
    \midrule
    {\multirow{4.2}{*}{\rotatebox[origin=c]{0}{CN $vs.$ MCI $vs.$ AD}}} & {\multirow{2}{*}{\rotatebox[origin=c]{0}{ADNI}}} & \textbf{$\mathcal{R}_{\text{stat}}$} & \text{0.7287}$\pm$\text{0.04} & \text{0.7239}$\pm$\text{0.03} & \textbf{0.7309}$\pm$\text{0.03}\\
    & & \textbf{Ours} & \textbf{0.7919}$\pm$\text{0.03} & \text{0.7847}$\pm$\text{0.03} & \text{0.7713}$\pm$\text{0.04}\\
    \cmidrule(lr){2-6}
    & {\multirow{2}{*}{\rotatebox[origin=c]{0}{GARD}}} & \textbf{$\mathcal{R}_{\text{stat}}$} & \text{0.7338}$\pm$\text{0.03} & \text{0.7338}$\pm$\text{0.03} & \textbf{0.7405}$\pm$\text{0.03}\\
    & & \textbf{Ours} & \textbf{0.7972}$\pm$\text{0.04} & \text{0.7858}$\pm$\text{0.04} & \text{0.7681}$\pm$\text{0.04}\\
    \bottomrule
    \end{tabular}}
    \end{table}
        
\subsubsection{Quantitative comparison with DL methods}
        \rev{As for the DL baselines, we adopted CNN-based~\cite{he2016deep,zhang2021explainable} and Transformer-based~\cite{dosovitskiy2021an,jang2022m3t, zhang2022diffusion} methods, including ResNet18~\cite{he2016deep}, ResAttNet~\cite{zhang2021explainable}, ViT~\cite{dosovitskiy2021an}, M3T~\cite{jang2022m3t}, and DSTANet~\cite{zhang2022diffusion}. ResNet18~\cite{he2016deep} is known as a CNN-based representative image classifier and a model that derives superior performance in various prediction tasks. ResAttNet~\cite{zhang2021explainable} is a variant of the ResNet that combines the internal self-attention modules. ViT~\cite{dosovitskiy2021an} is a series of Transformers that deals with image patches as a sequential input. M3T~\cite{jang2022m3t} synergically models the integration of CNN and Transformer architectures and exploits various 2D views (\ie, multi-planes and slices) from 3D MRIs. DSTANet~\cite{zhang2022diffusion} is designed by the Transformer that replaces dot-product attention to diffusion kernel attention while utilizing brain ROIs as an input sequence, similar to our LiCoL. In Table~\ref{tab:density_performance_with_cGM} and Table~\ref{tab: multi-cls}, the results of all DL baselines were also distilled with cGMs for a fair assessment, and the strategy of cGM augmentation resulted in notable performance improvements on all DL baselines when comparing to without augmenting the cGMs (refer to Table S16 and Table S18). Interestingly, comparative methods based on sequential modeling generally performed better in all scenarios, whereas ViT derived relatively lower performance than pure ResNet and other DL baselines. We conjectured such varying trends might be due to the lack of training samples, as ViT requires large samples for representation learning.}

        \rev{It is noteworthy that, albeit a lightweight classifier, LiCoL derived performance on par with DL baselines across all binary scenarios and achieved state-of-the-art performance in the multi-class scenario. In particular, DL models are built on layers of networks using complex non-linearity with numerous learning parameters, yielding high classification performance yet inevitably sacrificing interpretability. Contrarily, we designed the LiCoL with linearity so that the internal representations could be interpretable, such as other conventional ML models. Accordingly, the rationale behind any classification decision was straightforward to understand without additional tools or computations.}

\subsection{Discrepant ROIs analysis between AD-effect and statistical ROIs}\label{sec:discrepant-roi-analysis}
    \rev{We further investigated the variability in (m)AUC performance when using common or additive $\mathcal{R}_{\text{eff}}$ discrepant ROIs (\ie, $\mathcal{R}'_{\text{eff}}$) and $\mathcal{R}_{\text{stat}}$ discrepant ROIs (\ie, $\mathcal{R}'_{\text{stat}}$). To evaluate the validity of those ROIs, we separated the set of ROIs into three categories: (i) baseline ($\mathcal{R}_{\text{eff}}$ and $\mathcal{R}_{\text{stat}}$), (ii) intersection ($\mathcal{R}_{\text{eff}} \cap \mathcal{R}_{\text{stat}}$), and (iii) union ($\mathcal{R}_{\text{eff}} \cup \mathcal{R}'_{\text{stat}}$ or $\mathcal{R}_{\text{stat}} \cup \mathcal{R}'_{\text{eff}}$). Here, we re-trained our LiCoL as a classifier according to each ROI combination. As revealed in Table~\ref{tab:invest_discrepant_rois}, the best predictive performance was achieved by the model that used the baseline $\mathcal{R}_{\text{eff}}$. We also presented the results of using intersection and union derived from the $\mathcal{R}_{\text{eff}}$ (Ours), revealing a slightly decreasing tendency compared with the baseline in several scenarios. That is, $\mathcal{R}'_{\text{stat}}$ could be regarded as ROIs that would not significantly contribute to the improvement of predictive performance despite showing statistical significance. Meanwhile, the union with $\mathcal{R}_{\text{stat}}$ in all scenarios exhibited an average (m)AUC performance improvement (ADNI: 1.49\%$\uparrow$ and GARD: 1.11\%$\uparrow$) over baseline. We thus believe that utilizing discrepant ROIs $\mathcal{R}'_{\text{eff}}$ contributed to enhanced predictive scores as they only reflect the distinct differential regions among disease-related variations resulting from disease progression. In this light, $\mathcal{R}'_{\text{eff}}$ could be elucidated as potential biomarkers that have not appeared in the statistical test but were substantial.}

    \rev{From the comprehensive results, we investigated two instances over the $\mathcal{R}'_{\text{eff}}$ and intersection ROIs that were revealed across all scenarios of both ADNI and GARD. The intersection ROIs were L.PreCG, L/R.SFG, L/R.MFG, L.IFGtriang, L.REC, R.INS, L/R.HIP, L.IPG, R.SMG, R.ANG, L/R.STG, L/R.MTG, L/R.ITG, and R.ACCpre. Since those discovered regions are known as the most prevalent AD-related landmarks for AD progression~\cite{risacher2009baseline}, we are convinced that our $\mathcal{R}_{\text{eff}}$ was highlighted well and reflected prominent regions resulting from anatomical variations at each clinical stage. As $\mathcal{R}'_{\text{eff}}$, L.PFCventmed was observed across all scenarios for ADNI and the CN $vs.$ MCI and CN $vs.$ MCI $vs.$ AD scenarios for GARD, R.PFCventmed was discovered within the CN $vs.$ MCI and CN $vs.$ MCI $vs.$ AD scenarios for both ADNI and GARD. In ADNI, we observed that L.MCC, L.MOG, R.IPG, L.SMG, L/R.IFGoperc, and R.PCUN were shown as $\mathcal{R}'_{\text{eff}}$ in the CN $vs.$ MCI and CN $vs.$ MCI $vs.$ AD scenarios, whereas L.ACCpre was discovered in all scenarios. Moreover, L.ROL was found in the MCI $vs.$ AD and CN $vs.$ MCI $vs.$ AD scenarios, whereas L.SMA appeared in both the CN $vs.$ MCI and MCI $vs.$ AD scenarios, and L.ACCpre was identified in all scenarios. On GARD, we discovered L.PCUN in the CN $vs.$ MCI and CN $vs.$ MCI $vs.$ AD scenarios. To facilitate intuitive understanding, we adopted common and discrepant ROIs that appeared on ADNI and GARD and visualized them via 3D-volume mapping using longitudinal samples, as depicted in Fig. S4.}

    \begin{figure*}
    \centering
    \includegraphics[width=.96\textwidth]{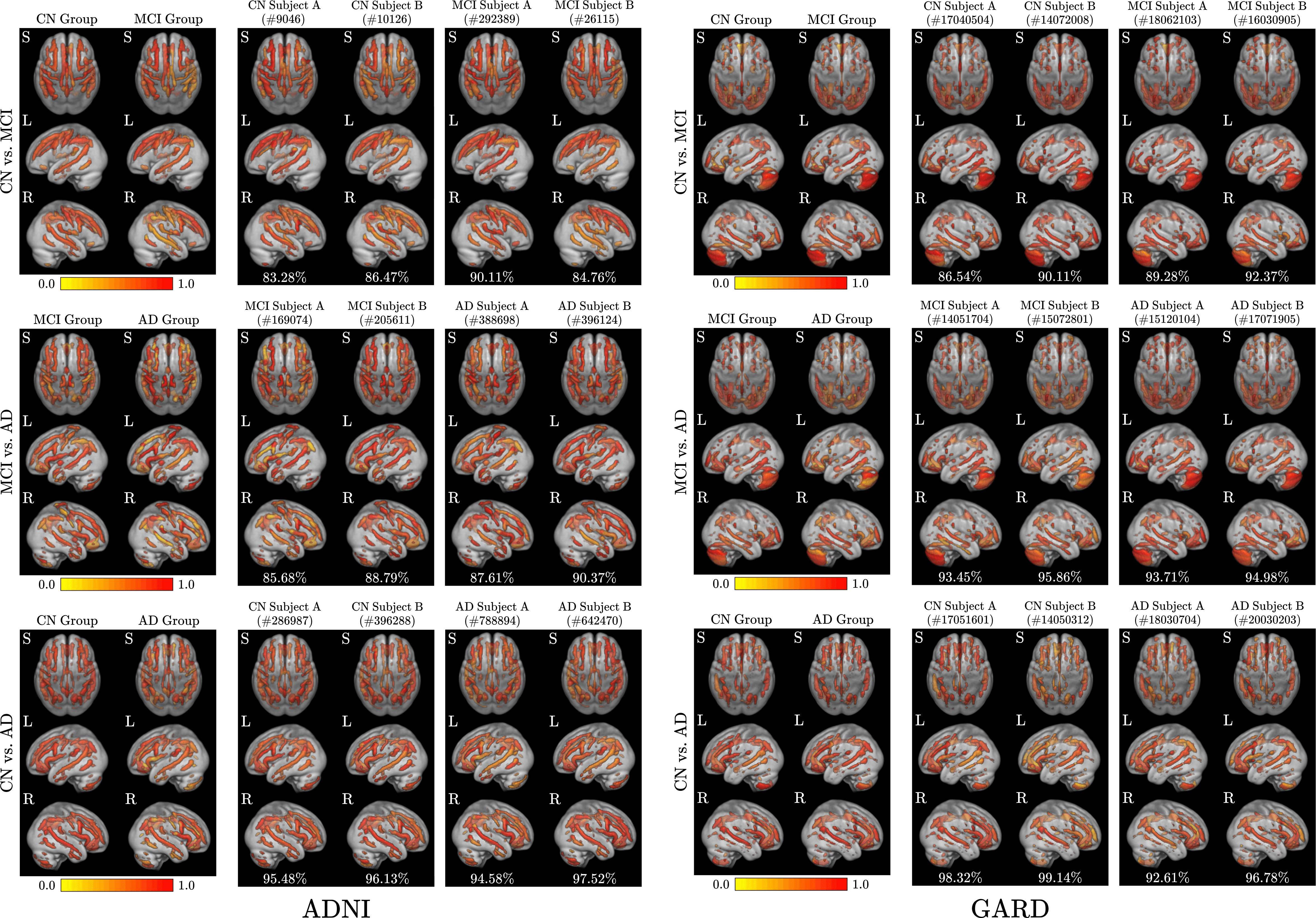} % 0.94
    \caption{Visualization of a normalized AD-relatedness index over the group-wise (first column) and individuals (second and third columns) on the ADNI and GARD datasets. Each row denotes a scenario, and the $\#$number on the top of the second and third columns indicates the image ID over randomly selected longitudinal samples. The percentage scores reported below in each scenario indicate the posterior probability (\ie, accuracy) of the respective longitudinal samples inferred by the trained LiCoL. Here, S, L, and R refer to the orientation of the sagittal plane, left, and right brain planes, respectively.}
    \label{fig:ADNI-GARD-ad-relatedness}
    \end{figure*}

\subsection{AD-relatedness indices for LiCoL’s interpretability}\label{exp:licol-interpretability}
    As we devised our LiCoL by revoking the incomprehensible non-linearities in mind, we attentively embraced a linear function that intuitively interprets the final decision. Accordingly, we present the mathematical development of the LiCoL as a linear function. 
    
    By exploiting the simplified function $\hat{\mathbf{y}} = \sigma\left(\operatorname{MP}_{\downarrow}\operatorname{MP}_{\rightarrow}\left(\mathbf{CA}\hat{\mathbf{x}}_s\right)c_2+c_1\right)$ in Supplementary H, we analyzed the input-dependent representation of $\mathbf{CA}\hat{\mathbf{x}}_s$ (termed AD-relatedness index) that revealed the contribution of each ROI for patient-specific prediction. We used the counterfactual-guided attention map $\mathbf{CA}$ as the region-wise AD-related importance to support the LiCoL decision in any classification scenario. \ks{In Fig.~\ref{fig:ADNI-GARD-ad-relatedness}, we illustrate the region-wise AD-relatedness indices over the group and individual patients in each scenario on ADNI and GARD datasets. For the group-wise investigation (first column in the ADNI and GARD results), we averaged the AD-relatedness indices for all test samples corresponding to the respective clinical stage. Among all scenarios related to the group-wise investigation, we observed that the majority of AD-effect ROIs for CN $vs.$ AD were closely related to the AD prediction (\ie, most red-colored regions). In this scenario, as the morphological difference \emph{w.r.t} brain atrophies were relatively immense, our LiCoL attended to the most ROIs with prominent GM density changes. Meanwhile, the AD-relatedness index for individual patients (second and third columns in the ADNI and GARD results) was randomly visualized among the arbitrary longitudinal subjects that were accurately classified with high confidence (\ie, accuracy). It was intriguing that each patient revealed a varying AD-related index despite being guided by a coherent AD-effect ROIs $\mathcal{R}_\text{eff}$. This showed that our LiCoL considered the regional differences in GM density between clinical groups and reflected the individual discriminative characteristics regarding the patient's status.}
    
    Overall, the ROIs of L/R.HIP, L/R.STG, L/R.IFGoperc, and L.PreCG exhibited especially high AD-relatedness indices among AD-effect ROIs in all scenarios. \ks{Particularly, the HIP is one of the prominent regions that shows atrophy in AD~\cite{foundas1997atrophy} and is treated as an index of AD neuropathology~\cite{gosche2002hippocampal}. This region is further related to the MCI~\cite{desikan2009automated} or the early marker~\cite{de1989early}. The temporal region composed of STG is known as a region where structural variations occur in the AD-manifested brain~\cite{davies1987quantitative}.} In CN $vs.$ MCI and CN $vs.$ AD scenarios, R.INS and L/R.PCUN did not exhibit as essential ROIs for CN prediction, whereas those ROIs appeared as crucial markers when used to diagnose MCI or AD patients~\cite{foundas1997atrophy,karas2004global}. Notably, L.SMA was an ROI that markedly contributed to diagnosing MCI, and R.IPG was found to have a significant ROI with a high AD-relatedness index in AD. \ks{Additionally, the regions of L/R.SFG and L/R.MFG, located in the frontal area, are also captured in all scenarios, which is associated with AD~\cite{davies1987quantitative}. Based on in-depth analyses via LiCoL's transparency, we clinically validated that the DL model (\ie, CMG) well-generated the target properties of which regions in the brain to change the classifier's prediction to the target class as the brain disease progressed. Thereby, AD-relatedness indices could provide neuroscientific insights for disease stratification in patient groups and understanding of individuals' symptoms, facilitating the direction of more specified diagnostic criteria for personalized inspection.}
    % Based on such in-depth analyses, we clinically validated that the DL model well-generated the target properties of which regions in the brain to change the classifier's prediction to the target class, as the brain disease progressed through quantitative analysis of counterfactual-guided deep features based on GM density.
    % We further analyzed the AD-relatedness index for each patient to provide the opportunity for more specified personal-diagnostic criteria. With such interpretability and a novel AD-relatedness index, our LiCoL can provide a quantitative and numerical interpretation of the anatomical variations in the brain, resulting from AD progression in each patient as well as a group of patients.

    \begin{table}[t]\footnotesize \setlength{\tabcolsep}{2.2pt}
    \caption{Examination of precision and recall according to the number of samples for each age group in the MCI $vs.$ AD scenario. These results can be certified to alleviate spurious correlations by additionally using the cGM. Here, the set of AD-effect ROIs (Ours) and statistical ROIs (baseline) is defined as $\mathcal{R}_\text{eff}$ and $\mathcal{R}_\text{stat}$, respectively, and the usage of augmented cGM is concisely indicated as w/ cGM.}
    \centering
    \scalebox{0.87}{
    \begin{tabular}{cccccccccc}
    \toprule
    \multicolumn{1}{c}{\multirow{2}{*}{\textbf{Data}}} & \multicolumn{1}{c}{\multirow{1.5}{*}{\textbf{Age}}} & \multicolumn{1}{c}{\multirow{2}{*}{\textbf{MCI}}} & \multicolumn{1}{c}{\multirow{2}{*}{\textbf{AD}}} & \multicolumn{2}{c}{\textbf{$\mathcal{R}_\text{stat}$}} & \multicolumn{2}{c}{\textbf{$\mathcal{R}_\text{stat}$ w/ cGM}} & \multicolumn{2}{c}{\textbf{$\mathcal{R}_\text{eff}$ (Ours)}}\\
    \cmidrule(lr){5-6} \cmidrule(lr){7-8} \cmidrule(lr){9-10}
    & \multirow{1}{*}{\textbf{(years)}} &  &  & \textbf{Precision} & \textbf{Recall} & \textbf{Precision} & \textbf{Recall} & \textbf{Precision} & \textbf{Recall} \\
    \midrule
    \parbox[t]{1.5mm}{\multirow{3}{*}{\rotatebox[origin=c]{90}{ADNI}}} & $60 \sim 70$ & 206 & 65 & \multicolumn{1}{c}{0.6897} &\multicolumn{1}{c}{0.8995} &\multicolumn{1}{c}{0.7871} &\multicolumn{1}{c}{0.9144} & \multicolumn{1}{c}{\textbf{0.8313}} & \multicolumn{1}{c}{0.9331} \\
     & $70 \sim 80$ & 359 & 174 & \multicolumn{1}{c}{0.8201} &\multicolumn{1}{c}{0.9067} &\multicolumn{1}{c}{0.8641} &\multicolumn{1}{c}{0.9356} & \multicolumn{1}{c}{\textbf{0.8951}} & \multicolumn{1}{c}{0.9415} \\
    & $80 \sim 90$ & 149 & 101 & \multicolumn{1}{c}{0.7273} &\multicolumn{1}{c}{0.9331} &\multicolumn{1}{c}{0.8318} &\multicolumn{1}{c}{0.9428} & \multicolumn{1}{c}{\textbf{0.8491}} & \multicolumn{1}{c}{0.9628} \\
    \midrule
    \parbox[t]{1.5mm}{\multirow{3}{*}{\rotatebox[origin=c]{90}{GARD}}} & $60 \sim 70$ & 130 & 19 & \multicolumn{1}{c}{0.6338} &\multicolumn{1}{c}{0.9339} &\multicolumn{1}{c}{0.7452} &\multicolumn{1}{c}{0.9773} & \multicolumn{1}{c}{\textbf{0.7780}} & \multicolumn{1}{c}{0.9840} \\
     & $70 \sim 80$ & 181 & 53 & \multicolumn{1}{c}{0.7016} &\multicolumn{1}{c}{0.9211} &\multicolumn{1}{c}{0.7838} &\multicolumn{1}{c}{0.9276} & \multicolumn{1}{c}{\textbf{0.8093}} & \multicolumn{1}{c}{0.9777} \\
    & $80 \sim 90$ & 50 & 34 & \multicolumn{1}{c}{0.5742} &\multicolumn{1}{c}{0.8942} &\multicolumn{1}{c}{0.7661} &\multicolumn{1}{c}{0.8997} & \multicolumn{1}{c}{\textbf{0.7957}} & \multicolumn{1}{c}{0.9146} \\
    \bottomrule
    \end{tabular}}
    \label{tab: capability of CF data}
    \end{table}

\subsection{Counterfactual images for spurious correlations}
    We further explored whether issues encountered in the medical field could be overcome using cGM augmentation. Indeed, accurate prevalence estimates of dementia by age, or other demographic information for different levels of severity could support effective medical treatment. Particularly, age constitutes the greatest of the various risk factors \emph{w.r.t} AD manifestation, with the percentage of the population who have AD increasing dramatically with age. However, although 3\% of people aged 65 $\sim$ 74, 17\% of people aged 75 $\sim$ 84, and 32\% of people aged 85 or older have been diagnosed as AD patients~\cite{hebert2013alzheimer}, it is important to note that AD is not a normal part of aging~\cite{nelson2011alzheimer}, and older age alone is not considered a sufficient factor to cause AD. Various medical studies~\cite{degrave2021ai,mahmood2021detecting} call this phenomenon spurious correlation, which occurs when two factors appear to be correlated to each other, but in fact are not~\cite{simon1954spurious}. Furthermore, few AD cases are identified when the patient's age is younger than 65 years, and further data on the prevalence of brain disease are scarce. Hence, samples of a specific age or clinical stage are lacking, which leads to an undesirable bias. We thus exploited the cGMs as augmented samples to alleviate these practical issues. For this purpose, we used the set of $\mathcal{R}_{\text{stat}}$ and $\mathcal{R}_{\text{eff}}$ as the classifier's input (use our LiCoL) to demonstrate the fidelity of cGM augmentation by showing that the performance shifts, which depend on whether augmented data are included in the training.

    As reported in Table~\ref{tab: capability of CF data}, the precision of the trained LiCoL that used $\mathcal{R}_{\text{stat}}$ was found to be relatively lower in young AD (\ie, aged 60 $\sim$ 70) on ADNI as well as elderly MCI (\ie, aged 80 $\sim$ 90) on GARD. Furthermore, unlike in the ADNI, a lack of samples appeared in the 80 $\sim$ 90 age range in GARD, and this age range exhibited the lowest precision. In this light, we can assume that precision degradation was derived due to the undesirable spurious correlation. However, when statistical ROIs were constructed along with cGM (\ie, $\mathcal{R}_\text{stat}$ w/ cGM), the performance with augmentation was noticeably improved across all age ranges. Particularly, it should be noted that the precision for the aged 80 $\sim$ 90 on GARD dramatically increased by +19.19\%. Through these results, we argued that the auxiliary cGMs of a patient at different clinical stages provided informative features, so that LiCoL could learn how to correctly distinguish the discriminative characteristics of each clinical stage, depending on the age range. Thus, we suggest distilling cGMs can be beneficial for mitigating the data-hungry problem in practical settings and promising alternatives for solving spurious correlations.

    \section{Discussion}\label{sec:discussion}
    The goal of GM density manipulation in this study is to normalize the individual brain image so that it aligns to a consistent anatomical space to account for personal variability. From this perspective, GM density maps might induce the disappearance of individual-specific shape information and minor structural distortions due to mapping into the shared template during the manipulation. However, contrary to these concerns, the variations in density values across the GM density maps after registration still represent crucial information. GM density maps take into account the fact that the size and shape of brains can vary considerably among individuals during the registration process; hence GM density values within these maps present \textit{GM volume} so that it enables us to facilitate meaningful comparisons by focusing on the quantity of GM density. Furthermore, these maps are sensitive to subtle variations in the GM volume or concentration, which may occur in the early stages of AD manifestation, even before morphological changes are visible on linear-registered MRI scans. Thanks to such advantages, it can better depict brain atrophies, a hallmark of Alzheimer's or other neurodegenerative diseases, making it widely used in medical image analysis as one of the quantitative features~\cite{karas2004global,karas2007precuneus,apostolova2007three}. Nonetheless, exploiting the entire GM density maps alone might be inadequate for AD classification as it is hard to capture the subtle structural variations because of the aligning process. To overcome this drawback, we further utilized the ROI strategy to focus on specific regions where the GM density differences occur depending on the disease progression. In particular, since our AD-effect ROIs are only constructed via \textit{GM density values} in highlighted regions that are quite sensitive to the variability of AD influence and highly related to AD manifestation, it is possible to boost the effectiveness of the ROI-based classification. To further verify the applicability of such in-depth analysis, an extensive evaluation was performed by utilizing other promising counterfactual-based approaches, as reported in Supplementary K.
    
    \ks{Our predictive evaluation showed that LiCoL, despite being a linear classifier, outperformed all the comparative ML-based models considered in our study with substantial margins and was comparable to DL-based models. Under the linear property, We have further demonstrated that our LiCoL possesses a monotonicity similar to ML-based models, which are prone to comprehend the internal working. In particular, through the AD-relatedness index induced by the LiCoL's inherent transparency, we could observe that each patient and group had a different contribution within ROIs, depending on the severity of morphological variations. By virtue of the LiCoL's superiority, we thus claim that our method could be a stepping stone toward a comprehensible explanation of the deep models' decisions and for predicting and analyzing neurodegenerative disease by leveraging quantitative figures.}
    % Our predictive evaluation showed that LiCoL, despite being a linear classifier, outperformed all the comparative ML-based models considered in our study with substantial margins and was comparable to DL-based models. Based on the AD-relatedness index induced by the interpretability of LiCoL, we further observed that each patient and group had a different contribution within ROIs, depending on the severity of morphological variations. This finding implies that attending the AD-effect ROIs to the input sample merely appropriately guides the LiCoL while considering each individual's unique characteristics, \ie, it does not have a dominant effect on classification. This finding implies that enforcing the AD-effect ROIs to individual patients effectively focuses on regions and reveals significant differences between clinical stages. In this way, we have further demonstrated that our LiCoL possesses a monotonicity similar to ML-based models, which are prone to comprehend the internal working while achieving superior performance. Consequently, we claim that our method could be a stepping stone toward a comprehensible explanation of the deep models' decisions and for predicting neurodegenerative disease from a clinician's perspective.
    
    % Thereby, the AD-relatedness index could provide neuroscientific insights for disease stratification in patient groups and understanding of individuals' symptoms such that it facilitates the direction of more specified diagnostic criteria for personalized inspection.

    \section{Conclusion}\label{sec:conclusion}
    In this work, we proposed a novel framework to analyze the counterfactual-induced visual explainability by transforming them into a GM density as a quantitative feature representation. Quantitative feature-based analysis using AD-effect ROIs helped conduct accurate numerical measurements based on the volumetric density changes in the brain caused by AD progression. Additionally, we designed a LiCoL, a simple shallow linear classifier, to boost the effectiveness of AD-effect ROIs while providing outstanding performance and the model's interpretability. Under LiCoL's transparency, we further produced an AD-relatedness index that can be identified to intuitively understand the AD-related landmarks for an individual subject and groups.

    \ks{We have viewed that utilizing visual explanation manipulated to quantitative figures is beneficial in yielding neuroscientific insights from a clinical aspect. However, there is still marginal room for improvement. Indeed, counterfactual reasoning regarding causality has to consider various confounders such as age, gender, or genetic factors, and the classification performance over multi-class scenarios must also be enhanced to provide more reliable interpretability. In this context, the future direction of this research would like to develop a model that adequately incorporates such demographic factors based on mutual relationships and derives outstanding performance in the multi-class scenarios while accounting for the nuanced differences among several disease stages via quantitative interpretation accordingly.}

% \appendices

% % you can choose not to have a title for an appendix
% % if you want by leaving the argument blank
% \section{}

\section*{Acknowledgement}
This work was supported by the Institute of Information \& communications Technology Planning \& Evaluation (IITP) grant funded by the Korea government (MSIT) No. 2022-0-00959 ((Part 2) Few-Shot Learning of Causal Inference in Vision and Language for Decision Making) and No. 2019-0-00079 (Department of Artificial Intelligence (Korea University)). This study was further supported by KBRI basic research program through Korea Brain Research Institute funded by the Ministry of Science and ICT (22-BR-03-05).

\bibliographystyle{IEEEtran}
\bibliography{main}

\begin{IEEEbiography}[{\includegraphics[width=1in,height=1.25in,clip,keepaspectratio]{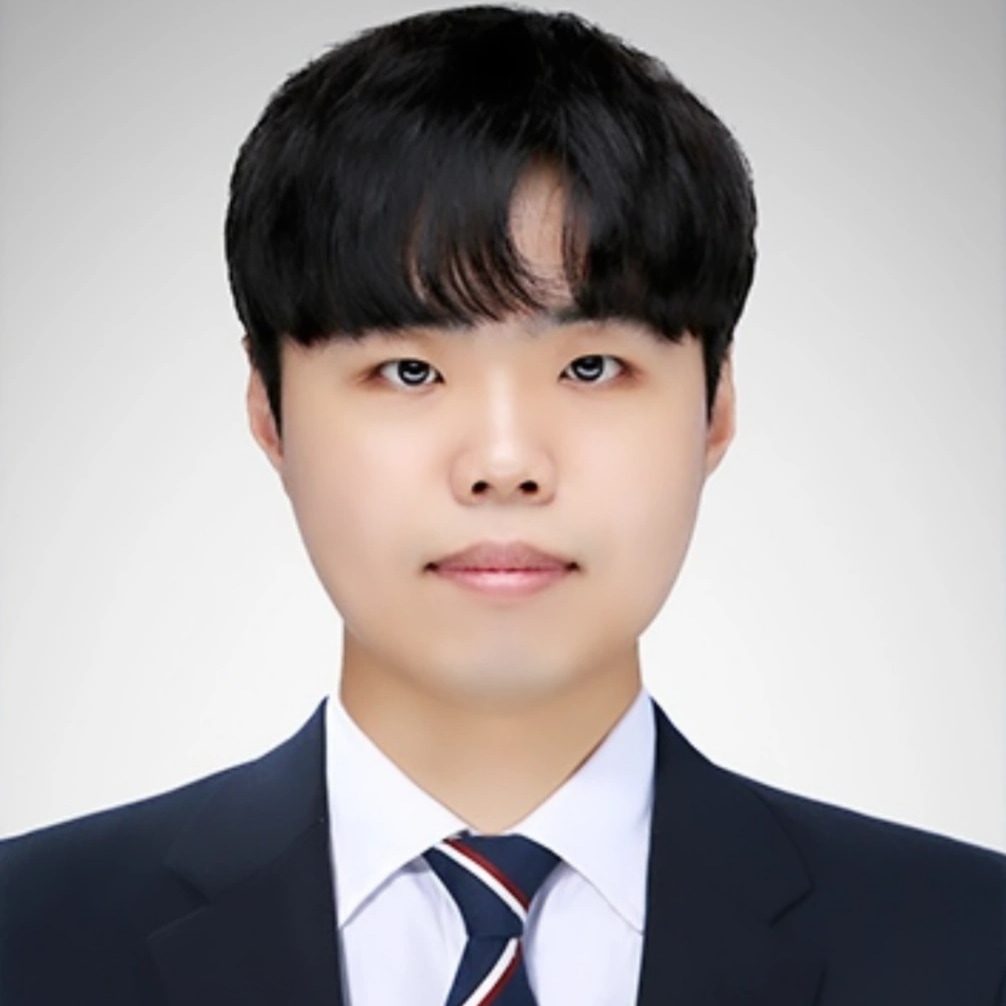}}]{Kwanseok Oh}
	received the B.S. degree in Electronic Control and Engineering from Hanbat National University, Daejeon, South Korea, in 2020. He is currently pursuing a Ph.D. degree with the Department of Artificial Intelligence, Korea University, Seoul, South Korea. His current research interests include explainable AI, computer vision, and machine/deep learning. 
\end{IEEEbiography}

\begin{IEEEbiography}[{\includegraphics[width=1in,height=1.1in,clip,keepaspectratio]{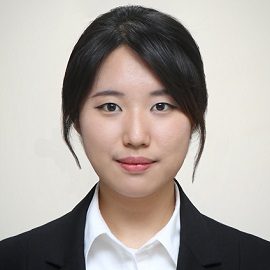}}]{Da-Woon Heo}
    received the M.S. degree in Brain and Cognitive Engineering from Korea University, Seoul, South Korea, in 2018. She is currently pursuing a Ph.D. degree with the Department of Artificial Intelligence, Korea University, Seoul, South Korea. Her current research interests include machine/deep learning, medical AI, and neuroscience.
\end{IEEEbiography}

\begin{IEEEbiography}[{\includegraphics[width=1in,height=1.1in,clip,keepaspectratio]{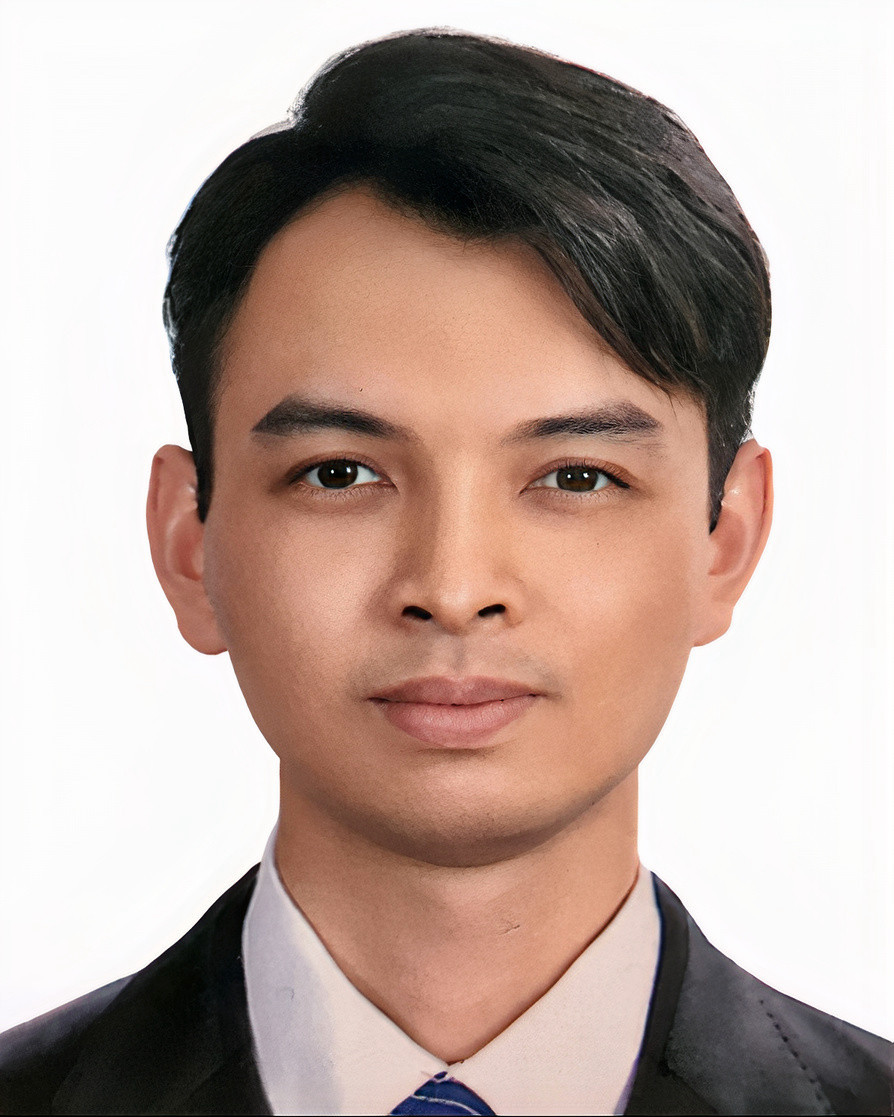}}]{Ahmad Wisnu Mulyadi}
    received the B.S. degree in Computer Science Education from the Indonesia University of Education, Bandung, Indonesia, in 2010. He is currently pursuing a Ph.D. degree with the Department of Brain and Cognitive Engineering, Korea University, Seoul, South Korea. His current research interests include machine/deep learning in healthcare, biomedical image analysis, and graph representation learning.
\end{IEEEbiography}

\begin{IEEEbiography}[{\includegraphics[width=1in,height=1.1in,clip,keepaspectratio]{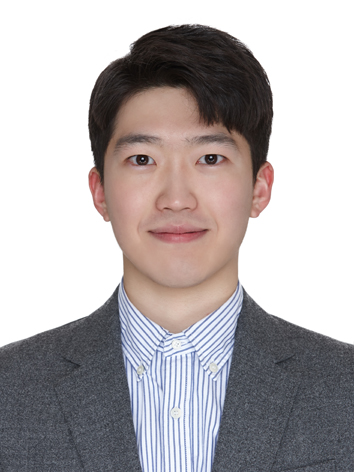}}]{Wonsik Jung}
    received the B.S. degree in Bio Medical Engineering from Konyang University, Daejeon, South Korea, in 2018. He is currently pursuing a Ph.D. degree with the Department of Brain and Cognitive Engineering, Korea University, Seoul, South Korea. His current research interests include computer vision, time-series modeling, and representation learning.
\end{IEEEbiography}

\begin{IEEEbiography}[{\includegraphics[width=1in,height=1.1in,clip,keepaspectratio]{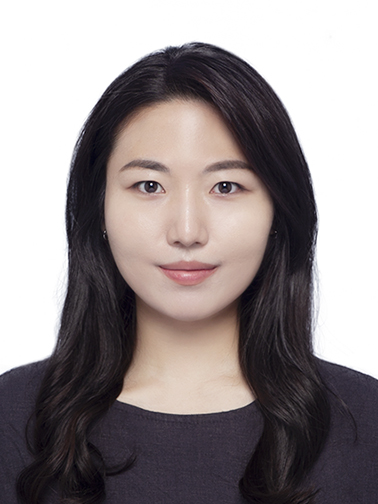}}]{Eunsong Kang}
    received the B.S. degree in Psychology from Korea University, Seoul, South Korea, in 2017. She is currently pursuing a Ph.D. degree with the Department of Brain and Cognitive Engineering, Korea University, Seoul, South Korea. Her current research interests include explainable AI, time-series modeling, and medical image analysis.
\end{IEEEbiography}

\begin{IEEEbiography}[{\includegraphics[width=1in,height=1.1in,clip,keepaspectratio]{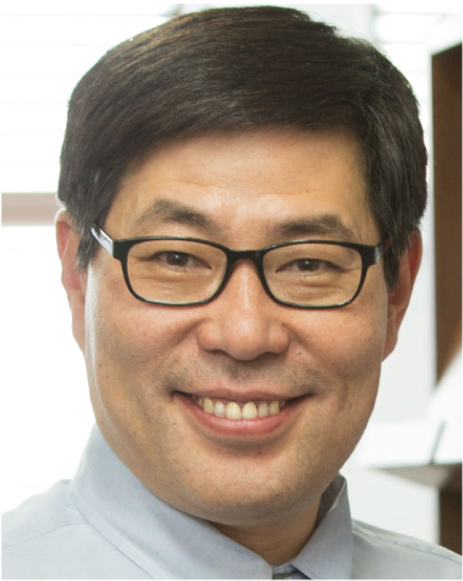}}]{Kun Ho Lee}
    received the B.S. degree from the Department of Genetic Engineering, Korea University, Seoul, Republic of Korea, in 1989, and the M.S. and Ph.D. degrees from the Department of Molecular Biology, Seoul National University, Seoul, in 1994 and 1998, respectively. He is currently an Associate Professor with the Department of Biomedical Science, Chosun University, Gwangju, Republic of Korea. He also works with the National Research Center for Dementia, Chosun University. His current research interests include brain image analysis and the development of prediction model for neurodegenerative diseases based on MRI and genetic variants.
\end{IEEEbiography}

% \begin{IEEEbiographynophoto}{Kun Ho Lee}

\begin{IEEEbiography}[{\includegraphics[width=1in,height=1.25in,clip,keepaspectratio]{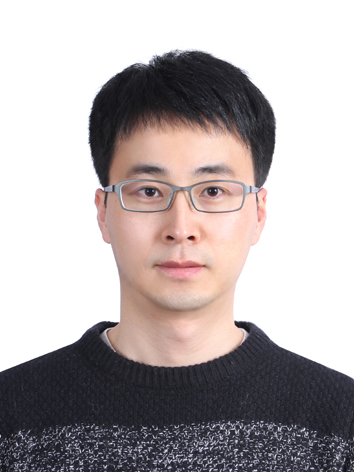}}]{Heung-Il Suk}
    received the B.S. and M.S. degrees in Computer Engineering from Pukyong National University, Busan, Korea, in 2004 and 2007, respectively, and the Ph.D. degree in Computer Science and Engineering from Korea University, Seoul, South Korea, in 2012.
	
    From 2012 to 2014, he was a Post-Doctoral Research Associate with the University of North Carolina at Chapel Hill, Chapel Hill, NC, USA. He is currently an Associate Professor at the Department of Artificial Intelligence and the Department of Brain and Cognitive Engineering, Korea University. He was awarded a Kakao Faculty Fellowship from Kakao and a Young Researcher Award from the Korean Society for Human Brain Mapping (KHBM) in 2018 and 2019, respectively.
    His research interests include machine/deep learning, explainable AI, biomedical data analysis, and brain-computer interface.

    Dr. Suk serves as an Editorial Board Member for Electronics, Frontiers in Neuroscience, Frontiers in Radiology, International Journal of Imaging Systems and Technology (IJIST), Clinical and Molecular Hepatology, and a Program Committee or a Reviewer for international conferences, including NeurIPS, ICML, ICLR, AAAI, IJCAI, CVPR, MICCAI, IPMI, MIDL, \etc
\end{IEEEbiography}


\end{document}